\documentclass[lettersize,journal]{IEEEtran}
\usepackage{amsmath,amsfonts}
\usepackage{algorithmic}
\usepackage{algorithm}
\usepackage{array}
\usepackage[caption=false,font=footnotesize,labelfont=rm,textfont=rm]{subfig}
\usepackage{textcomp}
\usepackage{stfloats}
\usepackage{url}
\usepackage{verbatim}
\usepackage{graphicx}
\usepackage{cite}
\usepackage[backref]{hyperref} 
\usepackage{times}
\usepackage{epsfig}
\usepackage{graphicx}
\usepackage{amsmath}
\usepackage{amssymb}
\usepackage[T1]{fontenc}
\usepackage[utf8]{inputenc}
\usepackage{authblk}
\usepackage{makecell}
\usepackage{float}
\usepackage{booktabs}
\usepackage{multirow}
\usepackage{arydshln}
\usepackage{bbding}
\usepackage{pifont}
\usepackage{color}
\usepackage{colortbl}
\usepackage[table]{xcolor}
\usepackage{mathrsfs}

\begin{document}

\title{What You See Is What You Detect: Towards better Object Densification in 3D detection}

\author{Tianran Liu, Zeping Zhang, Morteza Mousa Pasandi, Robert Laganiere
\thanks{This paper was supported, in part, by Synopsys under a partnership program of the Natural Sciences and Engineering Research Council of Canada (NSERC)}
\thanks{}
}

\markboth{}%
{Shell \MakeLowercase{\textit{et al.}}: A Sample Article Using IEEEtran.cls for IEEE Journals}


\maketitle

\begin{abstract}
    Recent works have demonstrated the importance of object completion in 3D Perception from Lidar signal. Several methods have been proposed in which modules were used to densify the point clouds produced by laser scanners, leading to better recall and more accurate results. Pursuing in that direction, we present, in this work, a counter-intuitive perspective: the widely-used full-shape completion approach actually lead to a higher error-upper bound especially for far away objects and small objects like pedestrians. Based on this observation, we introduce a visible part completion method that requires only 11.3\% of the prediction points that previous methods generate. To recover the dense representation, we propose a mesh-deformation-based method to augment the point set associated with visible foreground objects. Considering that our approach focuses only on visible part of the foreground objects to achieve accurate 3D detection, we named our method What You See Is What You Detect (WYSIWYD). Our proposed method is thus a detector-independent model that consists of 2 parts: an Intra-Frustum Segmentation Transformer (IFST) and a Mesh Depth Completion Network(MDCNet) that predicts the foreground depth from mesh deformation. This way, our model does not require the time-consuming full-depth completion task used by most pseudo-lidar-based methods. Our experimental evaluation shows that our approach can provide up to 12.2\% performance improvements over most of the public baseline models on the KITTI and NuScenes dataset bringing the state-of-the-art to a new level. The codes will be available at \textcolor[RGB]{0,0,255}{\url{{https://github.com/Orbis36/WYSIWYD}}}
\end{abstract}


\section{Introduction}

For high-performance autonomous driving perception, lidar is probably the most critical sensor. Although we have recently witnessed an increasing amount of work based on lidar-image multimodality input, the lidar features extractor is still used as the mainstream branch in most network design, since it provides accurate depth information.

The main limitation of lidar is the sparsity of its representations: with the depth increasing, the density of the geometric features obtained from the point cloud decreases rapidly. For the lidar-based methods, recent works \cite{xu2022behind, xu2021spg, wang2022sparse2dense, zhang2021pc, li2021sienet, shi2021graph, yuan2018pcn, qian2020end} have realized that better performance can be obtained by performing shape completion of objects with the network. However, it is important to point out that, most of these works are established under the premise that the first stage 3D Region of Interest(ROI) proposal is accurate enough, such that objects inside the proposed boxes can be refined and completed reliably. Considering that the Region Proposal Network(RPN) module at this stage still needs to face the sparsity of the lidar, especially for distant objects, this precondition can hardly be achieved. At the same time, even if the first stage detection fulfills the requirements, the completion task itself is inherently difficult for sparse inputs when relying only on lidar signal.

In early multimodal detection networks \cite{sindagi2019mvx, vora2020pointpainting, liang2019multi, xu2018pointfusion}, RGB features are used to decorate points or voxel features. With the research progressing, lifting 2D features to the pseudo lidar virtual points in 3D space has become increasingly popular. The success of recent cross-modality completion-based methods\cite{wu2023virtual, wu2022sparse, zhu2022vpfnet, jiao2023msmdfusion} confirms the appropriateness of this approach. Specifically, SFDNet\cite{wu2022sparse} has pioneered an attempt to introduce external points to complement the objects. This augmented data from depth completion, is equivalent to interpolating the lidar signals of the objects surface. By fusing this information with real lidar signals, complete objects can thus be generated from sparse point clouds. This enriched 3D point representation has led to significant improvements in performance.

\begin{figure}[t] 
	\centering 
	\includegraphics[width=0.48\textwidth]{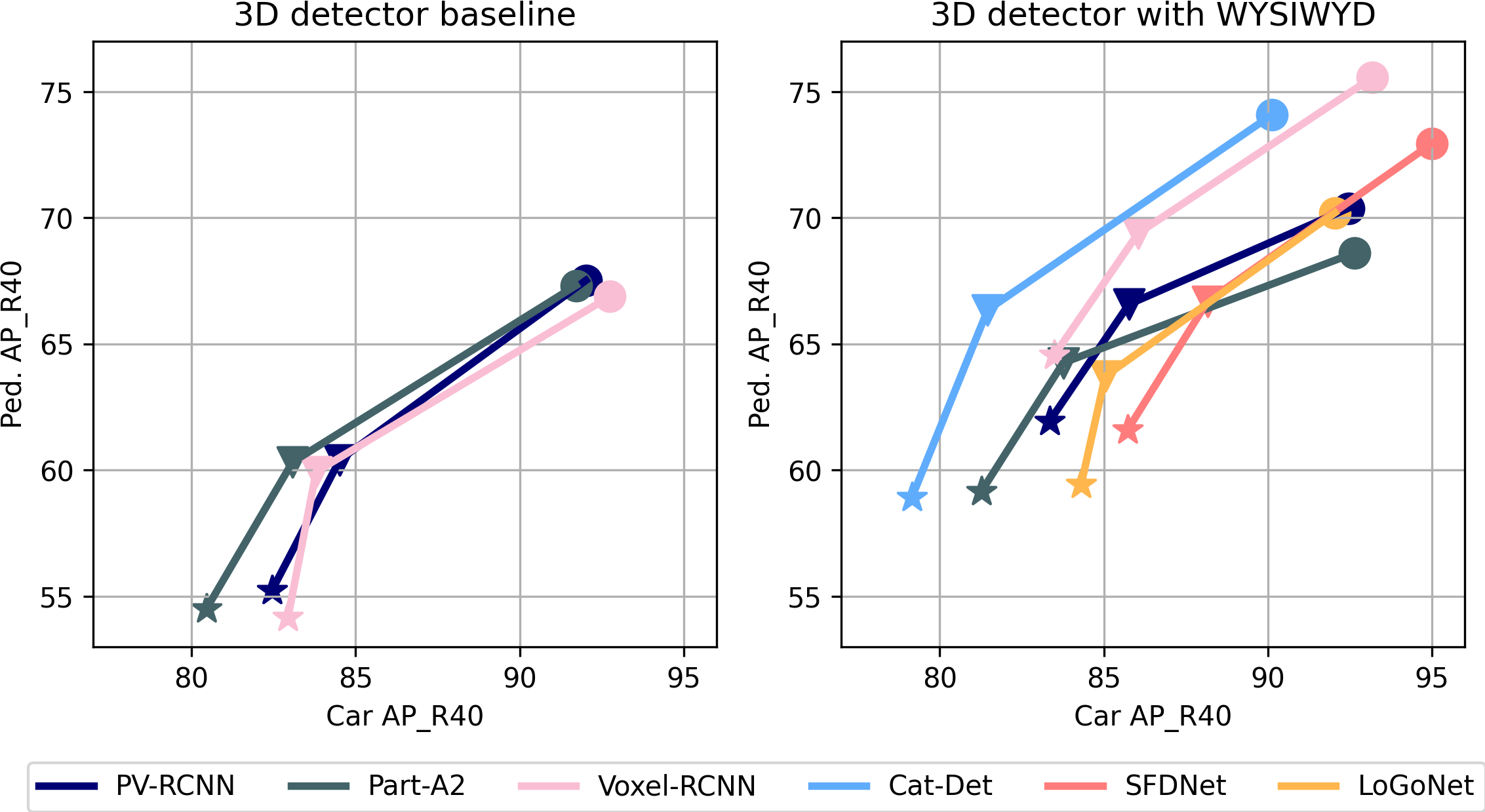} 
	\caption{Performance improvements brought by our proposed modules on KITTI 3D val. We compare the performance of the baseline model after combining WYSIWYD with that of today's SOTA solution in the right figure. Without 
		bells and whistles, by combining with our model, most baseline models can stand on the same level as today's SOTA solutions. Stars, triangles, and circles in the figure stand for detection under hard, moderate, and easy categories.} 
	\label{EvalOverall} 
\end{figure}

\begin{figure}[t]
	
	\centering
	\subfloat[Pseudo points (depth estimation) visualization from BEV]{
		\includegraphics[width=0.45\textwidth]{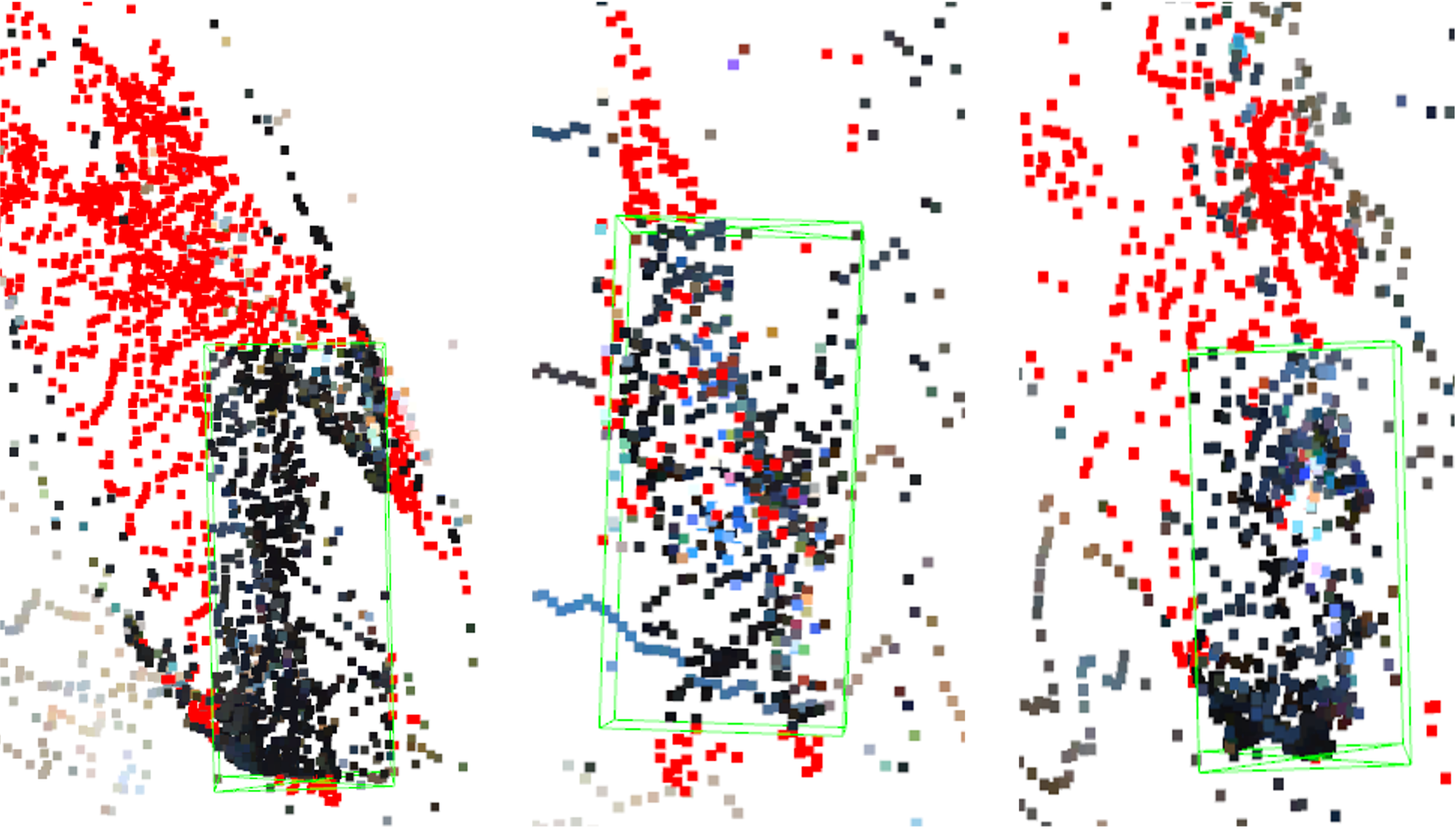}
	}
	\quad
	\subfloat[Visualization of error depth estimation from image plane]{
		\includegraphics[width=0.45\textwidth]{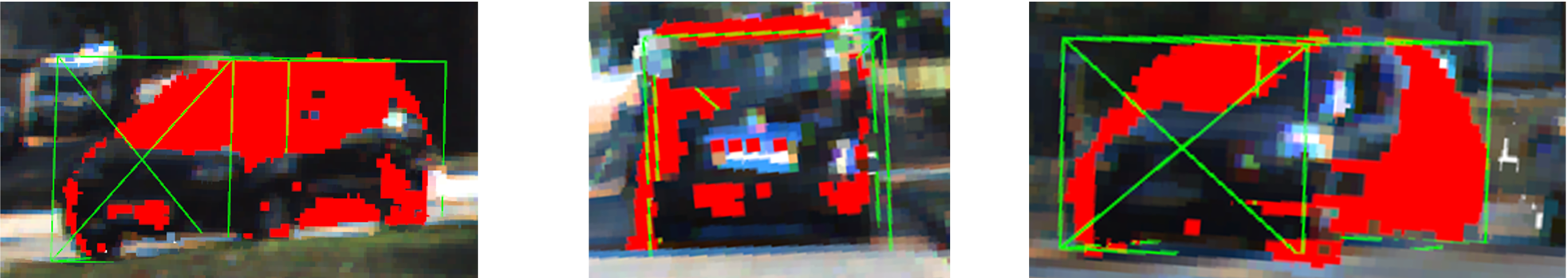}
	}
	\quad

	\caption{Boundary depth leak problem, ground truth 3D box shown in green, the estimated depth outside of the 3D box is marked in red. Most of the misprediction pixels occur along the boundary.}
	\label{boundary error}
	
\end{figure}

However, it is important to note that these pseudo-lidar representations are still plagued with inherent artifacts introduced by boundary depth dispersion problems, as shown in Fig \ref{boundary error}. Considering that the ground truth used to train the depth completion network is semi-dense and the inherent smoothing properties of convolutional layers used in these models, the depth estimation of boundary pixels always tends to leak toward the background. As a matter of fact, this issue seriously affects the accuracy of subsequent detections, as it will later be demonstrated in section \ref{fore-prove}.

Although a recent work \cite{wu2023virtual} has made it possible to reduce the inaccuracy of the depth estimation through a learnable discard module, models based on depth completion still suffer from two other issues. First, full frame depth completion itself is time-consuming and most of the generated background depth information is useless for the later process. A typical depth completion method like PENet\cite{hu2021penet} needs 161ms\cite{EPImple} per frame when CUDA synchronization is used. In consequence, a complete system based on such a pre-processing network will be far from real-time detection. Second, even if the outlier depth points can be dropped accurately, this discard-based strategy will actually further dilute the already limited amount of semantic information available for small objects.

In this work, we revisit the fundamental issues related to the quality of foreground depth and our objective is to make the entire detector to get rid of time-consuming full frame depth completion networks. Overall, we only complete the foreground points instead of performing global depth completion. This process can be divided into two steps: \textbf{foreground points segmentation} and \textbf{object densification}. 

Specifically, we demonstrate that visible part completion leads, in fact, to equivalent or even superior results than full shape completion used by most of the previous methods. This good performance also explains why methods like SFDNet\cite{wu2022sparse} or VirConv\cite{wu2023virtual} perform better than more traditional completion-based methods. To estimate the foreground region which needs to be densified, rather than relying on 3D RPNs that are insensitive to sparse objects, we choose to use the well-developed 2D instance segmentation networks. Within the 3D frustum of a specific segmented 2D object, considering the depth distribution of the point cloud, we design a lightweight Transformer, named IFST, to filter out the noise points.

Next, different from the existing models that use pseudo points from depth completion, here we chose to reconstruct the object directly from the lidar signal. The points from the IFST still contain some noise, so an ideal model should be able to control the shape and be invariant to noise as well. To fulfill such requirement, we integrated a mesh deformation approach to completion-based 3D detection for the first time. Through a lidar aggregation layer, we make the mesh learn the distribution of the lidar signal in a coarse to fine manner and restore the shape of the object progressively. Notably, the consistency of the resulting mesh is guaranteed and the vertex will not leak into the background benefiting from the specific guidance of the Laplacian loss.

It is also worth noting that although there are several works that claim to produce accurate point cloud completion\cite{yuan2018pcn, li2019pu}, they have all been tested in a noiseless or indoor environment. Few studies have demonstrated that they can be adapted to the case of sparse inputs. 

In summary, our contributions are as follows: 
\begin{itemize}
	\item By analyzing the previous depth pseudo point-based completion models, we propose to perform visible part completion, which has a higher detection upper bound from our verification experiments.
	\item We introduce a novel lightweight Intra-Frustum Segmentation Transformer that utilizes the 2D location prior and 3D locations to extract foreground points. As the main component, a mesh-deformation-based completion module is proposed to learn the visible shape of an object from the lidar signal.
	\item By combining our modules with the publicly available 3D detector baseline, extensive experiments have demonstrated that it is possible to provide up to 12.2\% performance improvement and obtain SOTA performance, especially in small object detection (see Figure \ref{EvalOverall}). 
\end{itemize}

\section{Related Work}

\textbf{Image perception guided 3D detection.}
In comparison to 3D detection, image-based detection, and segmentation tasks have reached a high level of performance in recent years. With the help of a 2D detection module, F-PointNet\cite{qi2018frustum} introduced intra-frustum detection to filter out the background points. F-Convnet\cite{wang2019frustum} then further developed the idea by proposing a sliding windows approach inside the frustum. A similar strategy was also adopted to obtain better fusion results in F-fusion\cite{zuo2021frustum}. Considering that distant objects can still be well-detected in RGB images, FarFrustum\cite{zhang2021faraway} further improved detection performance at different objects scale. The recent years have observed a rapid growth of works that utilize the well-developed 2D perception to guide 3D detection. FSF\cite{li2023fully} uses points in instance segmentation masks to augment the quality of lidar foreground queries before sending them to the Transformer Layer. MVF\cite{yin2021multimodal} utilizes the masks from 2D segmentation to add virtual points in 3D space. Frustumformer \cite{wang2023frustumformer} proposed an instance-aware resampling method to better utilize the more informative foreground pixels in BEV representation. 

\textbf{Fusion in homogeneous space.} From the accuracy point of view, the ROI (region of interest) level fusion performed in MV3D \cite{chen2017multi} and AVOD\cite{ku2018joint}, in which features are learned in separate spaces but fused by concatenation or other learning-based method directly, is undoubtedly not optimal. The Deep Continuous Fusion\cite{liang2018deep}, pioneered the exploration of interpolating RGB features and using them as subsidiary information for lidar voxels or points to participate in detection. This method can also be observed in several 3D detection pipelines \cite{vora2020pointpainting,xu2018pointfusion,sindagi2019mvx}. However, the mentioned cross-modality methods still adopt lidar as the mainstream detector and do not allow  point clouds to be complemented by the RGB features. The image information in the sparse part of the lidar remains underutilized. 

To address the inadequate information fusion due to point cloud sparsity, several recent approaches have realized that fusion performed in homogeneous space, normally 3D space, can significantly improve performance. Specifically, following the method in FCOS3D\cite{wang2021fcos3d}, the HomoFusion\cite{li2022homogeneous} projects lidar points to FOV and constructs depth confidence intervals for the RGB features, These points are sent to 3D space in order to perform homogeneous fusion. Furthermore, VPFNet\cite{zhu2022vpfnet} constructs virtual points from the foreground pixels to improve the utilization of local point clouds and RGB information. Through their virtual points multi-depth unprojection, MSMDFusion\cite{jiao2023msmdfusion} also reaches SOTA performance on nuScenes.

\textbf{Object Completion in 3D detection.}
Intuitively, a more complete shape can clearly improve detection accuracy. Considering that over half of hard samples in KITTI contain no more than 30 points, object completion in lidar representation has therefore a high potential to significantly improve the performance of a 3D detector. However, the estimation of shape and position requires strong prior knowledge, which is difficult to learn through detection networks. Current works in object completion-assisted detection can be broadly divided into two categories: using pseudo points obtained by pre-computed deep completion networks or through a sub-network densifying foreground points or voxels.

For the former, SFDNet\cite{wu2022sparse} introduced the first detection architecture that includes a depth completion module. VirConv\cite{wu2023virtual} further proposed Noise-Resistant Submanifold Convolution to identify and exclude the points for which the depth is incorrectly estimated. PseudoLidar++\cite{you2019pseudo} proposed a KNN-based traditional optimization algorithm, which corrects the position of pseudo points obtained by monocular 3D detection while maintaining the real-time nature of the network. 

For the latter, BtcDet \cite{xu2022behind} designed an occlusion prediction sub-network to recover the missing points by self or external occlusion in a cylindrical coordinate system. SPG\cite{xu2021spg} chose to densify the foreground points by an unsupervised expansion process.  Sparse2Dense\cite{wang2022sparse2dense} expressed this process more implicitly. In this work, densification is considered as a distance optimization problem under the hidden space.  PC-RGNN\cite{zhang2021pc} designed a GNN-based adversarial network and SieNet\cite{li2021sienet} proposed a PointNet-based interpolation method to generate the possible foreground points. GDCompletion\cite{shi2021graph} and PCN\cite{yuan2018pcn} proposed graph-based and point net-based methods to densify foreground points for improved downstream tasks. \cite{qian2020end} introduced a matching mechanism by calculating the voxel gradient to obtain a better location of missing points in foreground regions.

\begin{table*}
	
	\centering
	\setlength{\tabcolsep}{2.5mm}{
		\begin{tabular}{cccccccccccccc}
			\toprule[0.28mm]
			\multirow{2}{*}{\centering Methods}  & \multirow{2}{*}{\makecell[c]{Complete\\Category}} & \multicolumn{3}{c}{Car 3D $AP_{R40}$}& \multicolumn{3}{c}{Car BEV $AP_{R40}$} & \multicolumn{3}{c}{Ped. 3D $AP_{R40}$}& \multicolumn{3}{c}{Ped. BEV $AP_{R40}$}\cr
			\cmidrule[0.25mm](lr){3-5}\cmidrule[0.25mm](lr){6-8}\cmidrule[0.25mm](lr){9-11}\cmidrule[0.25mm](lr){12-14}
			& & Easy&Mod.&Hard&Easy&Mod.&Hard&Easy&Mod.&Hard&Easy&Mod.&Hard\cr
			\cmidrule[0.25mm](lr){1-14}

			\multirow{2}{*}{Point-RCNN\cite{shi2019pointrcnn}}& VP & \textbf{99.92}&\textbf{99.67}&\textbf{97.27}&\textbf{99.94}&\textbf{99.69}&\textbf{97.24}&\textbf{85.32}&\textbf{77.91}&68.25&\textbf{86.12}&\textbf{80.95}&\textbf{71.22} \\
			& FS & 97.27& 97.26& 97.19&97.29& 97.26& 97.19&77.86& 72.84& \textbf{70.35}&80.34& 74.89& 71.10\\
			\cmidrule[0.25mm](lr){1-14}
			
			\multirow{2}{*}{IA-SSD\cite{zhang2022not}}&\ VP & \textbf{99.92}& \textbf{99.77}&99.40&\textbf{99.91}&\textbf{99.27}&99.32&\textbf{75.58} &\textbf{71.38}&66.47&\textbf{77.53}& \textbf{75.11}& 70.37\\ 

			& FS & 99.71 & 99.47 & \textbf{99.42} & 99.65 & 99.22 & \textbf{99.38} & 71.92 & 69.79 & \textbf{68.20} & 74.89 & 74.37 & \textbf{72.86} \\
			\cmidrule[0.25mm](lr){1-14}
			\multirow{2}{*}{Voxel-RCNN \cite{deng2021voxel,chen2022focal}}& VP & 99.62&\textbf{99.70}&\textbf{99.62}&99.63&\textbf{99.72}&\textbf{99.67}&\textbf{91.94}&85.73&82.28&\textbf{93.86}&\textbf{89.61}&83.97\\
			
			& FS &  \textbf{99.96}&99.47&99.47&\textbf{99.99}&99.49&99.49&91.74&\textbf{88.64}&\textbf{85.47}&92.94& 89.58& \textbf{87.31}\\
			\cmidrule[0.25mm](lr){1-14}
			
			\multirow{2}{*}{PV-RCNN\cite{shi2020pv}}&\ VP & 99.47& 99.46&\textbf{99.46}&99.44&99.48&\textbf{99.47}& \textbf{88.74} &\textbf{84.09}&\textbf{79.20}&\textbf{92.64}& \textbf{88.00}& \textbf{83.15}\\ 
			& FS & \textbf{99.98} & \textbf{99.99} & 97.49 & \textbf{99.98} & \textbf{99.99} & 97.49 & 80.40 & 78.97 & 77.45 & 84.68 & 82.96 & 83.10 \\ 
			
			\bottomrule
	\end{tabular}}
	\vspace{2mm}
	\caption{The upper boundary performance (i.e. based on ground-truth shape completion)of different object completion forms a comparison on KITTI val: models mentioned in the table are typical point-based, voxel-based, and cross-aggregation-based methods. VP/FS: visible part completion or full shape completion. The bolded line is the method with the highest accuracy. Detection thresholds are 0.7/0.5/0.5 for Car and 0.5/0.5/0.5 for Pedestrian, The highest accuracy in the last 20 epochs are selected to list in the table}
	
\end{table*}
\section{Problem Fomulation}

As discussed, 3D detection approaches based on point completion have had great success. In this section, we will outline some of the core reasons leading to this success. Based on this, a better completion method and the steps to generate the ground truth for training will be introduced.
\vspace{-0.3cm}
\subsection{Hypothesis} \label{fore-prove}
A typical pseudo-points-based detector method needs an upstream depth completion module, in which the depth of every pixel is estimated. Here we use SFDNet\cite{wu2022sparse} as the testbed to explore the relation between the accuracy of foreground points and 3D detection. The authors of SFDNet mentioned PENet\cite{hu2021penet} and TWISE\cite{imran2021depth} in their paper and implementation but only adopted the latter in the published code. Consequently, we first simply replace this completion module with different depth completion methods to verify their effectiveness inside the downstream 3D detection. 

\begin{figure}[t] 
	\centering 
	\includegraphics[width=0.5\textwidth]{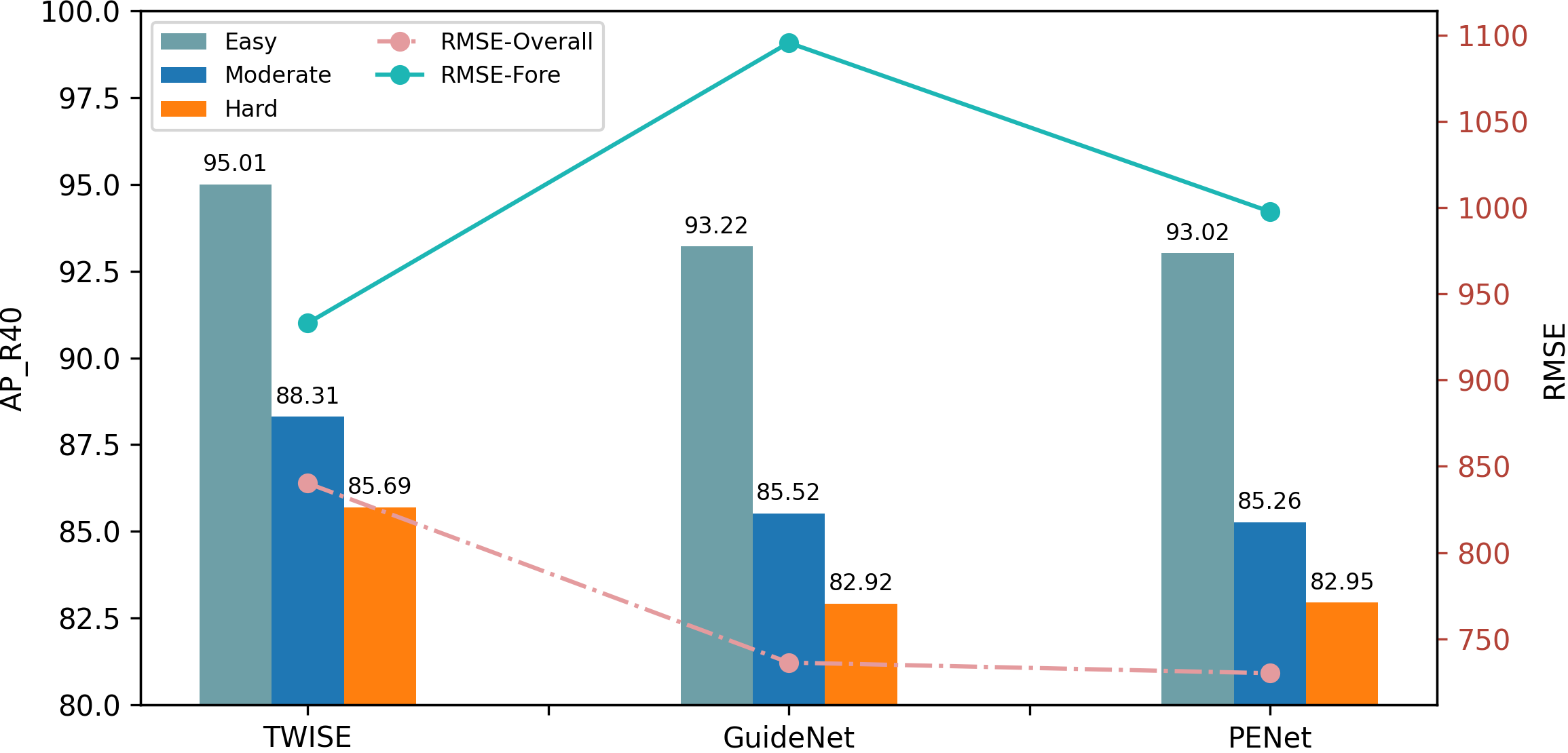} 
	\caption{SFDNet 3D AP with different depth completion methods. TWISE has the highest overall RMSE (dotted orange line) but the lowest RMSE for foreground objects as shown by plain green line. Methods to calculate foreground RMSE are shown in section  \uppercase\expandafter{\romannumeral3}-B. } 
	\label{RMSE} 
\end{figure}

As shown in Fig. \ref{RMSE}, the highest 3D detection performance occurs when TWISE is used. But when applying RMSE, the criterion for evaluating the performance of depth completion, TWISE obtains the worst results (as shown by the pink line). Considering that the RMSE evaluates both foreground and background depth, while the foreground is more critical to 3D detection, we speculate that TWISE\cite{imran2021depth} obtain a better performance in foreground depth prediction. To verify this hypothesis, we need to generate the ground truth of foreground depth, since the lidar points cannot guaranteed that all pixels in the 2D foreground region can get a depth estimate. This generation process is introduced in section \ref{p32} and will later be used to estimate the RMSE of the foreground objects in section \ref{comp-obj}.

\subsection{Dense depth generation of visible part} \label{p32}

With the annotated 3D boundary boxes and 2D masks\cite{heylen2021monocinis}, the points that can be projected to the specific mask of objects in the KITTI dataset can be extracted from the lidar file, denoted by $P=\left\{ p_1, p_2,  \cdots, p_Q \right\}$ $ p_i \in \mathbb{R}^{(n_i \times 3)}$, Q objects in total, $n_i$ points for the object i. The masks of all objects are denoted as $\mathcal{O}=\left\{ \Omega_1, \Omega_2,  \cdots, \Omega_Q \right\}$ $ \Omega_i \in \mathbb{N}^{(m_i \times 2)}$, $m_i$ pixels for the mask i. Considering the correspondence relationship, set P and $\mathcal{O}$ are equinumerous. We now want to get a dense depth that can make $\forall i$, $n_i$ = $m_i$. 

To generate a groundtruth for the dense visible part depth, we first get the full shape of the object and then obtain the depth of its visible part. For the KITTI dataset, we first maintain a pool of objects that contain every object $p_i$ extracted by 3D boxes, if $n_i > 20$ for cars and 10 for pedestrians. Secondly, for every object $p_i$ in the pool, we mirror it to recover the backside of the object, and the output obtained is denoted by A. Then we basically follow the heuristic $\mathcal{H}(A, B)$ mentioned in BtcDet\cite{xu2022behind} to find the best match B in the pool for the sample A. However, we noticed that this method can hardly reflect the shape of a real 2D mask when we project them to the image, especially when the object is very sparse. So here, the heuristics are modified as the follows: for every sample A, a best matching B will minimize the $\mathcal{G}(A, B)$ as shown in Equation \ref{eq1}. Here, we add a term to calculate the pixel-wise IOU between ground-truth and 2D mask obtained from the projection of completed objects. $\epsilon$ is an indicator function, when $n_i > 10$ , $\epsilon$ equal to zero otherwise 1. The $ \Gamma$ is a surjection between space location and 2D location: $\Gamma : \mathbb{R}^{(n_i \times 3)} \rightarrow \mathbb{N}^{(n_i \times 2)}$, determined by the intrinsic matrix. 
\begin{equation}\label{eq1}
	\mathcal{G}(A, B) = \mathcal{H}(A, B) + \epsilon IOU(c_A, \Gamma(p_B)) 
\end{equation}
\begin{equation}
	p^{\prime}_{A}=p_A + p_B \;\in \mathbb{R}^{(n_A + n_B) \times 3}
\end{equation}

\begin{figure}[htbp] 
	\centering 
	\includegraphics[width=0.5\textwidth]{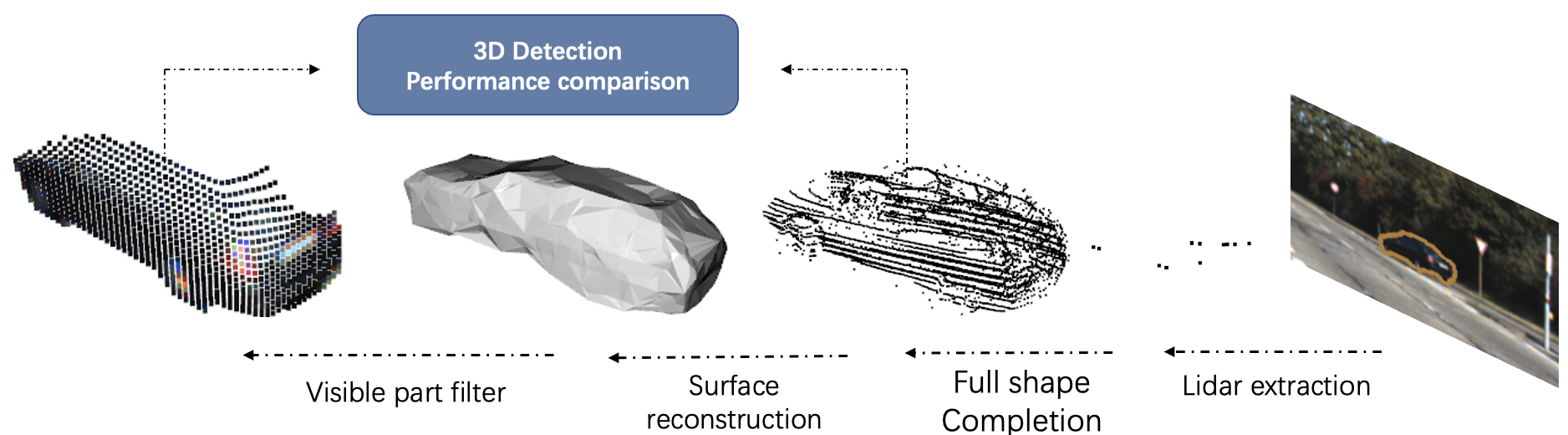} 
	\caption{From sparse lidar to visible part ground truth. With the full shape completion and visible part completion groundtruth, we can compare the upper boundary of detection performance (see Table 1).} 
	\label{Completion} 
\end{figure}

After the completed $p^{\prime}_{A}$ is obtained, a triangle mesh surface reconstruction denoted as $\mathcal{F}$, is adopted to get the hull of objects, denoted by $\mathcal{M}_{A}$. 
\begin{equation}
\mathcal{M}_{A}=\mathcal{F}(p^{\prime}_{A}) 
\end{equation}
Placing the obtained hull in 3D space, with the known camera intrinsic matrix, extrinsic matrix, and the pixel coordinates of mask $\Omega_i$, a ray casting model can be established. Note that since the rays from the pixel on the boundary sometimes misses the object, a volume expansion coefficient $\alpha$ is used here to make sure there is always a bijection from pixel set to depth set. For each pixel in the mask, when we project it to 3D space, there is a well-determined point $c_o: (0, y_c, z_c)$ and its direction vector $(\vec{x_c}, \vec{y_c}, \vec{z_c})$.
Then the depth of a specific pixel, denoted by $d_c$, can be obtained by:
\begin{equation}
L_c : \left(\begin{array}{l}
x \\
y \\
z
\end{array}\right)=\left(\begin{array}{l}
0 \\
y_c \\
z_c
\end{array}\right)+\lambda\left(\begin{array}{l}
\vec{x_c} \\
\vec{y_c} \\
\vec{z_c}
\end{array}\right)
\end{equation}
\begin{equation}
d_c = \left|L_c \cap \alpha \mathcal{M}_{A} -c_o \right| \cdot \left|\vec{x_c}\right|
\end{equation}

We can get the dense depth for the visible part after this process, as shown in Figure \ref{Completion}. With the ground truth obtained, the foreground RMSE for 3 different models have been calculated as shown in Figure \ref{RMSE}. The plain green line here demonstrates that among all candidate models, TWISE can achieve the best performance in foreground depth prediction. We thus showed that the key reason TWISE benefit downstream 3D detection is its high quality foreground depth prediction. In other words, the higher the foreground pseudo points quality, the better our 3D model performance. Pursuing with this idea, next we will present the result of using the generated ground truth to train a model directly. We will explore the upper boundary performance of this completion method and compare it with that obtained from the full completion.

\subsection{Completion Method Comparasion} \label{comp-obj}

Intuitively, a more complete object should lead to better results. However, from our experiments, the visible part of ground truth provides a counter-intuitive answer. 

Here we select 4 recent lidar detection models, which take the original lidar points and augmented points from different methods as input. As shown in Table 1, a wide performance increment can be observed especially in pedestrian detection. With only 11.3\% (on average) of the points contained in full shape completion, in most cases, the improvement of our proposed completion approach can be up to 8.5\%. We also noticed that worse results happen in the hard category, which can be explained by the small number of pixels available. 

This experiments proves that our proposed visual partial completion is in fact more suitable for 3D detection tasks. We believe this result can also be used to explain why the recent proposed depth completion model can outstand the vanilla shape completion-based methods: the former actually provides an overall higher upper boundary and needs less points to be predicted.

After having demonstrated the higher potential of the proposed visable part completion approach, our next question is if visable part foreground points matter, then how can we learn from these densified points and complete the objects? In the following section, we will introduce the proposed network which uses the points in frustum as input and generate pixel-wise visible part depth in a mesh-deformation manner.

\section{Proposed Method}

\subsection{Overview}
In order to identify the foreground points that need to be densified, we first project all lidar points onto a 2D mask which is obtained from an image instance segmentor. Although this approach successfully removes most irrelevant background points, it is worth mentioning that there may still be some noisy lidar points within the frustum due to inaccuracies in 2D segmentation and potential occlusions in 3D space. So we propose a lightweight transformer network to identify the foreground lidar points. In the subsequent step, by considering the depth of each pixel as a vertex, a mesh deformation-based network will densify the sparse lidar signal, allowing downstream detectors to benefit from this augmented pseudo-point representation. The overall structure of our model is illustrated in Figure \ref{IFST} and Figure \ref{FOP}.

\vspace{-0.3cm}
\subsection{Intra-Frustum points segmentation}
\begin{figure}[H] 
	\centering 
	\includegraphics[width=0.5\textwidth]{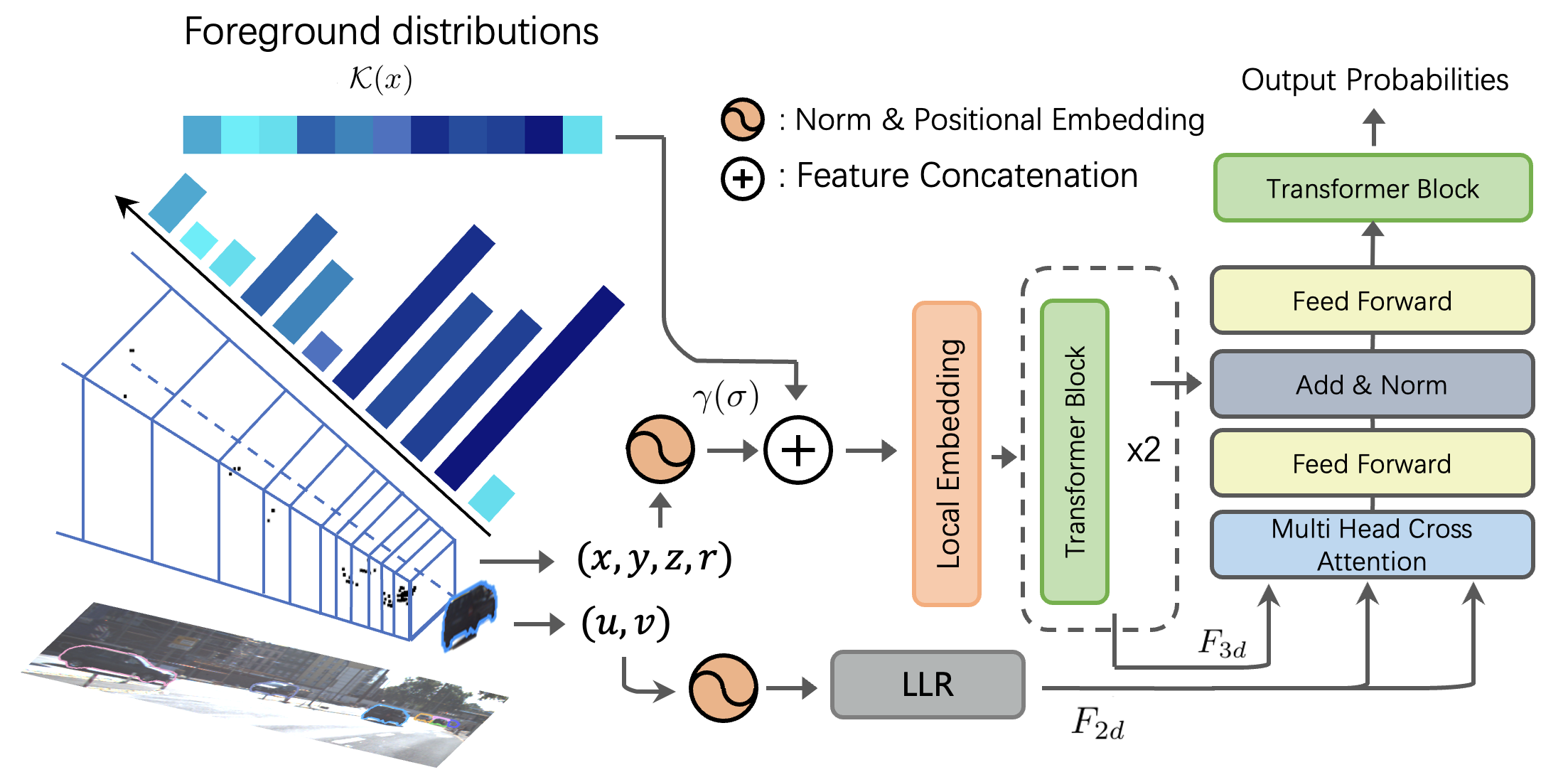} 
	\caption{Illustration of the IFST design, a light weight Intra-Frustum Transformer. Both size of the mask and 2d relative location are used as prior knowledge to guide the 3D points segmentation. } 
	\label{IFST} 
\end{figure}

\begin{figure*}[htbp] 
	\centering 
	\includegraphics[width=\textwidth]{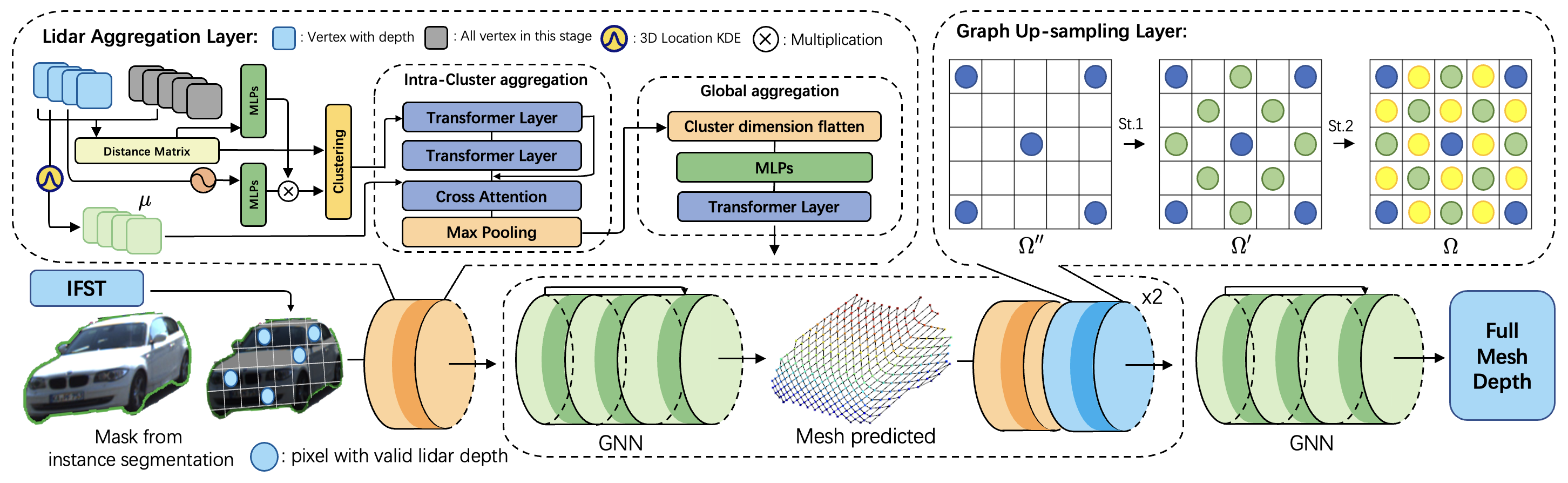} 
	\caption{The design of MDCNet: After obtain foreground points from IFST, the relative distance between depth empty vertex(pixel without depth) and valid vertex used to weight the depth by direct mutiplocation after embedding. With the relative distance, an intra-cluster aggregation and global aggregation followed to estimate the depth representation of specific pixels better. The GNN next used to propagate the features among vertex in the specific stage. Every pixel obtain the estimated depth after 3 stage up-sampling.} 
	\label{FOP} 
\end{figure*}

Large-scale point cloud segmentation tasks have been well developed in recent years\cite{qi2017pointnet++, lai2022stratified, guo2021pct}, however, few works focus on Intra-Frustum segmentation. From our experiments, we identify three essential characteristics that the performance of models can benefit from when conducting Intra-Frustum points segmentation.
\begin{itemize}
	\item \textbf{Eliminate downsampling operation:} Different from the traditional full scene point segmentation scenario, in a typical frustum produced by an image mask, each point has the potential to bring in useful semantic information. Considering that the overall number of points in frustums is only of the order of 100-1000, there is no need to adopt any downsampling strategy in the design of this network.
	\item \textbf{Guidance from 2D location:} During the projection from 3D to 2D, the background points naturally have a higher probability of being located on pixels at the boundary of the mask. This is a priori assumption has to be considered in order to obtain a more accurate point segmentation.
	\item \textbf{Guidance from Perspective relationship and Points Density:} Another geometric property that is often overlooked is that the objects in the image are naturally larger when close to the camera. This perspective phenomenon allows us to easily filter out some of the noise points. 
\end{itemize}
To optimize the utilization of the mentioned characteristics without compromising inference speed, we propose the Intra-Frustum Segmentation Transformer (IFST), as depicted in Figure \ref{IFST}. 

To use point density to guide the intra-frustum segmentation, we first divide the whole frustum by the density-adaptive splitting scheme shown in Algorithm 1. 

\begin{algorithm} 
	\caption{Density-adaptive splitting frustum} 
	\label{alg3} 
	\begin{algorithmic}[1]
		\REQUIRE The depth values of the n points associated with a foreground object are $p_x=\left\lbrace p_{x0}, p_{x1}, ...,p_{xn} \right\rbrace $. The elements of the set have been sorted from smallest to largest. Bandwidth h for density estimation and number of bins H.
		\STATE $\forall p_{xi}, \hat{f}_n(p_{xi})=\left| \frac{1}{n h} \sum_{j=0}^n \frac{1}{\sqrt{2 \pi}} \exp\left(-\frac{(p_{xi}-p_{xj})^2}{2h^2}\right) \right|$
		\STATE $f_{score}(p_{xi}) = H * Softmax(\hat{f}_n(p_{xi})^{-\frac{1}{2}})$
		\STATE $d_{bin} \leftarrow \varnothing, \quad j, k, d_{init} \leftarrow 0$
		\REPEAT
		\REPEAT 
		\STATE $d_{init} \leftarrow d_{init} + p_{xj}; \quad j \leftarrow j + 1$ 
		\UNTIL $\sum_{k}^{j}f_{score}(p_{xj}) = 1$
		\STATE $d_{bin} \leftarrow d_{bin} \cap \left\lbrace d_{init}\right\rbrace ; \quad k \leftarrow j$
		\UNTIL $\left| d_{bin}\right| = H$
	\end{algorithmic}
\end{algorithm}

By combining Gaussian kernel density estimation and softmax function, we assign to each point a score proportional to the density of points around it. By accumulating this value in the order of depth, we can give bins a finer granularity when the points are dense and vice versa. After partitioning the frustum space, a pointnet-like structure will process points in different sub-frustum and concatenate the feature of the mask size later. These sub-frustum wise feature will predicts the $\mathcal{K}(x)$ which shows the probability of each sub-frustum being foreground. This probability distribution will later be concatenated with 3D location embedding, and then sent to the subsequent network.

To get a better representation, we adopted two tricks from \cite{mildenhall2021nerf, lai2022stratified}, in the design of IFST. First, we transform the 2D/3D points into the frequency domain using sinusoidal functions as shown in equation 6. This projection allows similar inputs under Euclidean space to be clearly recognized by the network. $\sigma_i$ here represent the feature in the i-th dims and $\sum$ represent feature stack operation. Specifically, both $p_i$ and its projection $(u_i, v_i)$ are processed by this function separately, and the computed 3D embedding $\gamma(\sigma)$ will be concatenated with $\mathcal{K}(x)$ as shown in Fig 5. Secondly, to better describe the local features of the point cloud, an SA-Layer\cite{qi2017pointnet++} is used to aggregate the features of local neighbors before the 3D features are processed by a transformer layer.

\begin{equation}
\begin{aligned}
\gamma(\sigma)&=\sum_{i=0}^{5}\left[\sin \left(2^0 \pi \sigma_i\right), \cos \left(2^0 \pi \sigma_i\right), \cdots, \right. \\& \left. \sin \left(2^{L-1} \pi \sigma_i\right), \cos \left(2^{L-1} \pi \sigma_i\right)\right]
\end{aligned}
\end{equation}
The 2D location will be processed by several stacks of Linear-LayerNorm-Relu layers, denoted by LLR in Figure 5, to get $F_{2d}$
. $F_{3d}$, the feature from 3D stream will then be guided by $F_{2d}$ in a cross-attention manner. More details on this attention pipeline will be introduced in the section \ref{experiment_sec}. The final output is the probability for each point to be part of foreground. The role of the IFST is therefore to filter out the noise/background points on the objects identified by the instance segmentation module.
\subsection{Mesh deformation based foreground depth prediction}

The existing depth completion models always suffer from boundary depth dispersion problem, which is actually due to the tendency of convolutional networks to smooth the signal. Here we regard the pixel for which depth needs to be estimated as the vertices of a deformable mesh in 3D space. 

The specific network structure of this module is described in Figure 4. Inspired by recently proposed depth completion networks\cite{lin2022dynamic, zhang2023completionformer}, the core of our Mesh Depth Completion Network (MDCNet) design is a geometric position-based hierarchical Transformer that allows the model to learn from points at different locations while still maintaining a strong prior: to estimate the depth of a specific pixel, neighboring lidar points in 2D space are more referential. It is the role of the aggregation layer to make the network learn from real lidar point distribution and densify the mesh in an iterative manner.

Based on the previous work\cite{wang2018pixel2mesh, 9878407}, we also adopt a coarse-to-fine strategy to get the final shape. Given the 2D location of mask region $\Omega \in \mathbb{N}^{m \times 2}$, we first downsample the dense pixels to $\frac{1}{2}$ and then $\frac{1}{5}$ of the original by using the reverse process of the graph up-sampling layer presented in Figure \ref{FOP}, to get $\Omega^\prime$ and $\Omega^{\prime\prime}$, $\left| \Omega^{\prime\prime} \right| = 0.2m$. With the points $p^{\prime} \in \mathbb{N}^{t \times 3}$ filtered by IFST, their pixels location $\Omega_L \in \mathbb{N}^{t \times 2}$ can be calculated by the camera internal and external parameters. $\Omega^{L}$ should be a subset of $\Omega$, with $\left| \Omega^{L} \cap \Omega^{\prime\prime}\right| \geq 0$. Let the depths of these points denoted by $p^{\prime}_d \in \mathbb{R^+}^{t \times 1}$, in this (first) stage, the depth of pixels $\Omega^{\prime\prime} - \Omega^{L}$ need to be estimated by the obtained $p^{\prime}_d$ and $\Omega_L$.

Here, we propose an explicit local-to-global feature aggregation strategy to estimate the depth of the pixels on a mask. In general, given a specific pixel using $\Omega^{\prime\prime}_i$ to represent its 2D location, we first sort the t pixels locations $\Omega_L$ by $\left\| \Omega^{\prime\prime}_i - \Omega_L \right\|_2$, i.e. the Euclidean distance. We then estimate the KDE embedding $\mu$ for $p \prime$, which will later used to guide the aggregation. Next, we calculate the distance matrix between different pixels with and without lidar depth and process them with a MLP. This embedded 2D distance will be multiplied by the lidar features to get the relative location-weighted 3D features . This process can be described by Equation 7. $\mathcal{F}_{b} \in \mathbb{R}^{m \times t \times c}$, c is features dimension. Further, using the distance matrix, we divide the features into $\eta$ chunks, denoted by $\mathcal{F}_{bi}\in \mathbb{R}^{m \times \frac{1}{\eta}t \times c}$, to explore the intra-cluster relationship.
\begin{equation}
\mathcal{F}_{b} = MLPs(Embed(p^{\prime}_d)) * MLPs(\left\| \Omega^{\prime\prime} - \Omega_{L} \right\|_2)
\end{equation}
For features in different chunks,  we apply 2 layers of transformer encoder along the first dimension of $\mathcal{F}_{b}$ as shown in Equation 8. The stack operation denoted by $\sum$. This process allows the network to find a better representation for lidar in different distance bins. $\mathcal{F}_{bi}$ will then be used as a query for the attention matrix calculated by the KDE feature of corresponding lidar points.$\oplus$ in equation 8 stands for matrix multiplication.
\begin{equation}
\begin{aligned}
	\mathcal{F}_{bi} &= \sum_{j=0}^{s} SelfAtt(\mathcal{F}_{bij}) \quad \mathcal{F}_{bij} \in \mathbb{R}^{1 \times \frac{1}{\eta}t \times c} \\
	\mathcal{F}_{bi} &= sigmoid(MLPs(\mu_i) \oplus MLPs(\mu_i)) \oplus \mathcal{F}_{bi}
\end{aligned}
\end{equation}
At the end of cluster-wise aggregation, we dilute the intra-cluster feature for each vertex by max pooling to get new $\mathcal{F}_{bi} \in \mathbb{R}^{m \times \frac{1}{\eta} \times c} \rightarrow \mathcal{F}^{\prime}_{bi} \in \mathbb{R}^{m \times 1 \times c}$, so for all $\eta$ cluster, the feature before global aggregation is $\mathcal{F}^{\prime}_{b} \in \mathbb{R}^{m \times \eta \times c}$

For global aggregation, after the flattening in the last 2 dimensions of $\mathcal{F}^{\prime}_{b}$, the dimension of representation for a different vertex is $\left| \eta * c \right|$, which corresponds to the feature learnt from all clusters. The final transformer block was added to allow the network to learn features from other vertex directly, instead of by multiple neighborhood propagation in later GNNs.
Note that the hop of our network is quite larger than the classic scenario for GNNs, e.g. a social network or recommender systems, in which hop is around 6-10\cite{dou2020enhancing, kipf2016semi}. In our scenario, the distance from vertex to vertex may need over 200 hops (proportional to the number of pixels in the mask), this design actually provides a short-cut for the vertex to exchange features.

The following GNNs aggregate and pass the features to the neighbor of every vertex. In our design, to maintain the stability of gradient flow, and for every 2 layers of GNNs, we add a residual connection. We leverage spectrum-free graph convolutions following \cite{9878407}. Given the feature on vertex \textbf{f} and its neighbor $\mathcal{N}(i)$, the specific design is shown in Equation 9.
\begin{equation}
\mathbf{f}^\prime_i = \frac{1}{1+|\mathcal{N}(i)|} \left[\mathbf{W}_0 \mathbf{f}_i+\mathbf{b}_0+\sum_{j \in \mathcal{N}(i)}\left(\mathbf{W}_1 \mathbf{f}_j+\mathbf{b}_1\right)\right]
\end{equation}
where $\mathbf{W_0}$ and $\mathbf{W_1}$ are learnable parameters for the vertex itself and its neighbours. After 6 GNN layers, a regression head is used to predict the depth of every vertex on the mesh, i.e. the $\Omega^{\prime\prime}$. This process will be iteratively repeated 3 times with $\Omega^{\prime\prime}$, then $\Omega^{\prime}$ and $\Omega$ to obtain a dense depth for the object. 
\subsection{Training Losses}

The losses of the proposed modules can be divided into 2 parts, loss for segmentation and loss for mesh regression. Specifically, lidar segmentation is here a binary classification task, and a simple BCE Loss is adopted to provide guidance as shown in Equation 10, with $y_i$ being denote the label of $p_i$.
\begin{equation}
L_{seg}=\frac{1}{n}\sum_{i=0}^{n}y_i \cdot \log \sigma\left(x_i\right)+\left(1-y_i\right) \cdot \log \left(1-\sigma\left(x_i\right)\right)
\end{equation}
The mesh regression loss is composed of the location loss and mesh shape loss. We combine the MSE losses in all different stages as the location loss as shown in the first term of $L_{mesh}$ in Equation 11. The $\lambda_i$ in the early stage will be higher. For the shape loss, N represents the number of vertices in the mesh. The second and third items in $L_{mesh}$ aim to control the length of the edges in the predicted mesh and provide consistency among the normals of adjacent faces. This approach effectively prevents the occurrence of a long tail problem in the estimated points. The $loc$ denotes the predicted 3D depth of a specific vertex, and $n_i$ represents the normal vector of the triangular plane on the mesh. To balance the different loss terms, we introduce $\omega_1$, $\omega_2$, and $\lambda_m$.
\begin{equation}\label{loss}
\begin{aligned}
	L_{mesh} &= \sum_{i=1}^{3}\lambda_iMSE(loc, \hat{loc}) + \omega_1L_{edge} + \omega_2L_{con} \\
	L_{con} &= \frac{1}{N}\sum_{i=0}^{N}1 - cos(n_i, n_j), \quad j=Neighbour(i) \\
	L_{edge} &= \frac{1}{N}\sum_{i=0}^{N}\left\| loc_i, loc_j\right\|_2 , \quad j=Neighbour(i) 
\end{aligned} 
\end{equation}

\begin{equation}
	L = L_{seg} + \lambda_mL_{mesh}
\end{equation}

\section{Experiments} \label{experiment_sec}

\begin{table*}[h]
	
	\centering
	\setlength{\tabcolsep}{1.43mm}{
		\begin{tabular}{cccccccccccccccc}
			\toprule[0.25mm]
			\toprule[0.25mm]
			\multirow{2}{*}{\centering Methods}  & \multirow{2}{*}{\makecell[c]{Reference}} & \multirow{2}{*}{\makecell[c]{Modality}}& \multirow{2}{*}{\makecell[c]{With \\ WYSIWYD}} & \multicolumn{3}{c}{Car 3D $AP_{R40}$}& \multicolumn{3}{c}{Car BEV $AP_{R40}$} & \multicolumn{3}{c}{Ped. 3D $AP_{R40}$}& \multicolumn{3}{c}{Ped. BEV $AP_{R40}$}\cr
			\cmidrule[0.25mm](lr){5-7}\cmidrule[0.25mm](lr){8-10}\cmidrule[0.25mm](lr){11-13}\cmidrule[0.25mm](lr){14-16}
			& & & &Easy&Mod.&Hard&Easy&Mod.&Hard&Easy&Mod.&Hard&Easy&Mod.&Hard\cr
			\cmidrule[0.25mm](lr){1-16}
			
			\multirow{2}{*}{CenterPoint\cite{yin2021center}} & \multirow{2}{*}{CVPR2021}& L & \ding{55}& 89.83 &78.87& 75.79&92.35&87.86&85.26&48.63&46.18&42.09&53.42&51.32&47.81 \\
			& & L+I & \checkmark & 90.48 & 79.60 & 76.93 & 94.14 & 88.97 & 86.65 & 66.53 & 62.11 & 58.07 & 73.02 & 69.68 & 64.83 \\
			\cmidrule[0.25mm](lr){1-16}
			
			\multirow{2}{*}{Point-RCNN\cite{shi2019pointrcnn}} & \multirow{2}{*}{CVPR2019}& L & \ding{55}& 91.99 &80.26& 78.04&92.93&87.79&84.63&65.96&57.98&49.86&69.00&60.92&52.80 \\
			& & L+I & \checkmark &91.23&80.86&74.12&95.63&87.94&83.13&74.34&65.37&56.48&78.01&68.88&59.81 \\
			\cmidrule[0.25mm](lr){1-16}
			
			\multirow{2}{*}{PV-RCNN\cite{shi2020pv}}& \multirow{2}{*}{CVPR2020}& L & \ding{55}&92.02&84.52&82.45&92.94&90.74&88.59&67.52&60.41&55.23&69.76&63.49&58.85 \\
			& & L+I &\checkmark  & 92.45&85.76&83.34&95.58&91.68&89.64&70.38&66.57&61.93&75.43&71.02&66.23 \\
			\cmidrule[0.25mm](lr){1-16}
			
			\multirow{2}{*}{Voxel-RCNN \cite{chen2022focal}}& \multirow{2}{*}{AAAI2021}& L &\ding{55}  & 92.75&85.30&82.94&95.80&91.35&88.99&66.88&59.94&54.16&69.62&63.02&58.02 \\
			& & L+I &\checkmark  & 92.60&85.84&83.43&95.64&91.82&89.58&\textcolor[rgb]{1, 0, 0}{\textbf{75.56}}&\textcolor[rgb]{1, 0, 0}{\textbf{69.38}}&\textcolor[rgb]{1, 0, 0}{\textbf{64.56}}&\textcolor[rgb]{1, 0, 0}{\textbf{80.10}}&\textcolor[rgb]{1, 0, 0}{\textbf{75.25}}&\textcolor[rgb]{1, 0, 0}{\textbf{68.70}} \\
			\cmidrule[0.25mm](lr){1-16}

			\multirow{2}{*}{Part-$A^2$\cite{shi2020points}}& \multirow{2}{*}{TPAMI2020}& L & \ding{55}& 91.72&83.08&80.45&94.51&90.41&88.23&67.32&60.32&54.49&69.80&63.10&58.15 \\
			& & L+I &\checkmark  & 92.64&83.76&81.27&\textcolor[rgb]{1, 0, 0}{\textbf{95.94}}&89.86&89.22&68.61&64.25&59.16&75.13&70.54&65.38 \\
			
			\hdashline
			\hdashline
			\specialrule{0em}{2pt}{2pt}
			
			CAT-Det \cite{zhang2022cat}& CVPR2022& L+I& N/A & 90.12&81.46&79.15&-&-&-&74.08&66.35&58.92&-&-&-\\	
			\cmidrule[0.25mm](lr){1-16}
						
			\multirow{1}{*}{SFDNet\cite{wu2022sparse}}& \multirow{1}{*}{CVPR2022}& L+I& N/A & \textbf{94.99}&88.16&85.72&95.80&91.80&91.41&72.94&66.69&61.59&75.64&69.71&64.70 \\
			\cmidrule[0.25mm](lr){1-16}
			
			\multirow{1}{*}{VFF-PVRCNN\cite{li2022voxel}}& \multirow{1}{*}{CVPR2022} & L+I& N/A & 92.55&85.54&83.09&95.37&91.33&90.74&72.18&65.01&60.11&77.01&69.39&64.72 \\
			\cmidrule[0.25mm](lr){1-16}
			\multirow{1}{*}{VirConv-T\cite{wu2023virtual}} & CVPR2023& L+I& N/A & 94.98&\textbf{89.82}&\textbf{88.01}&95.42&\textbf{93.82}&\textbf{91.60}&73.32&66.93&60.38&73.32&66.93&60.38 \\
			\cmidrule[0.25mm](lr){1-16}
			\multirow{1}{*}{LoGoNet\cite{li2023logonet}}  & CVPR2023 & L+I& N/A & 92.04&85.04&84.31&93.08&90.79&90.55&70.20&63.72&59.44&74.29&66.93&63.70 \\
			\bottomrule[0.25mm]
			\bottomrule[0.25mm]
			
	\end{tabular}}
	\vspace{2mm}
	
	\caption{Illustration of performance improvement brought by WYSIWYD on baseline models and the comparison with state-of-the-art solutions on \textbf{KITTI validation set}. The best results from augmented baseline models and SOTA methods are shown in bold. L: Lidar, L+I: Lidar and image. The best performance in different categories is marked in red. Here we use GT sampling in all baseline models but cancel this augmentation when combine them with WYSIWYD. we retrained all listed models if the code is available.}
	\label{ResultKitti}
	
\end{table*}
In this section, the experimental setup and related details are first introduced. Then we give a comparison between baseline models combined with WYSIWYD and previous SOTA solutions on both KITTI and nuScenes. Our code has been developed using the OpenPCDet toolbox\cite{openpcdet2020}.

\begin{table*}[h]
	
	\centering
	\setlength{\tabcolsep}{2.4mm}{
		\begin{tabular}{ccccccccccc}
			\toprule[0.25mm]
			\toprule[0.25mm]
			Methods  & Reference & Modality& With WYSIWYD & mAP$\uparrow$ & NDS$\uparrow$ & mATE$\downarrow$ & mASE$\downarrow$ & mAOE$\downarrow$ & mAVE$\downarrow$ & mAAE$\downarrow$ \cr
			
			\cmidrule[0.25mm](lr){1-11}
			
			\multirow{2}{*}{CenterPoint\cite{yin2021center}} & \multirow{2}{*}{CVPR2021}& L & \ding{55}& 0.860 &0.834&0.155&0.214&0.266&0.192&0.138 \\
			& & L+I & \checkmark & 0.875&0.842&0.149&0.211&0.269&\textbf{0.191}&\textbf{0.136} \\
			
			\cmidrule[0.25mm](lr){1-11}
			\multirow{2}{*}{SECOND\cite{yan2018second}} & \multirow{2}{*}{SENSORS2018}& L & \ding{55}& 0.813&0.785&0.165&0.215&0.223&0.454&0.154 \\
			& & L+I & \checkmark & 0.836&0.811&0.157&0.215&0.241&0.309&0.153 \\
			
			\cmidrule[0.25mm](lr){1-11}
			\multirow{2}{*}{VoxelNext\cite{chen2023voxenext}} & \multirow{2}{*}{CVPR2023}& L & \ding{55}& 0.860&0.832&0.158&0.212&0.273&0.189&0.144 \\
			& & L+I & \checkmark & \textcolor[rgb]{1, 0, 0}{\textbf{0.892}}&\textcolor[rgb]{1, 0, 0}{\textbf{0.852}}&\textcolor[rgb]{1, 0, 0}{\textbf{0.140}}&0.210&\textcolor[rgb]{1, 0, 0}{\textbf{0.231}}&0.211&0.144 \\
			
			\hdashline
			\hdashline
			\specialrule{0em}{2pt}{2pt}
			
			TransFusion \cite{bai2022transfusion}& CVPR2022& L+I& N/A & 0.891&0.850&0.146&\textbf{0.208}&0.232&0.214&0.149\\	
			\cmidrule[0.25mm](lr){1-11}			
			BEVFusion \cite{liu2023bevfusion}& ICRA2023& L+I& N/A & 0.885&0.849&0.146&0.216&0.223&0.205&0.144\\	
			
			\bottomrule[0.25mm]
			\bottomrule[0.25mm]
			
	\end{tabular}}
	\vspace{2mm}
	\caption{Performance Comparsion on \textbf{nuScences validation set}. Only the accuracy of pedestrians and cars is accounted for here. Best result are shown in bold. $\uparrow$ higher is better, $\downarrow$ lower is better.}
	\label{ResultNu}
\end{table*}

\subsection{Experimental setup}

\textbf{Dataset and ground truth generation} he KITTI\cite{geiger2012we} 3D object benchmark is one of the most famous datasets in autonomous driving perception. We follow the setting in previous works that split the training part into 3712 and 3768 samples as training and validation sets. In the following content, most experiments will be reported on the KITTI validation set. Compared to the former, nuScenes\cite{caesar2020nuscenes} is a benchmark dataset of a larger scale, which provides ten times more training data than KITTI in the form of continuous frame labeling. The performance of 3D detectors augmented with WYSIWYD generated pseudo points are tested on these 2 datasets. 

For the generation of the visible part ground truth, note that the nuScenes only provide a 2D instance mask for nuImage, so we pretrained the segmentation model on nuImage, and performed inference on nuScenes. Since these masks are not accurate enough, our MDCNet has only been trained on the KITTI visible part ground truth. We then performed a zero-shot inference to complete the objects in nuScenes. 

In addition to the above-mentioned scheme of using 2D mask labels to generate visible part ground truth, we also use masks from E2EC\cite{zhang2022e2ec} predictions.

The input of IFST is the points filtered by 2D masks, however, if we only use label masks in the training process, the noisy points will be very sparse. In consequence, when we infer the model on the mask provided by e2ec in real cases, the noise points can hardly be identified. Note that in our model training. we partly use masks from e2ec as long as the IOU between e2ec's predicted mask and the true mask is greater than 0.7.

\textbf{Evaluation metrics}
For the KITTI part we report results using average precision under 40 recall thresholds and 0.7, 0.5 IOU thresholds for cars and pedestrians respectively. The accuracy for lidar segmentation are measured by mIOU, which is calcualted from TP (True Positive), FP (False Positive) and FN (False Negative) as shown in Equation \ref{meanIOU}, c is the number of category. Here $c=2$, since we only split foreground from background points. 
\begin{equation}\label{meanIOU}
meanIOU = \frac{1}{C} \sum_{c=1}^C \frac{\mathrm{TP}_c}{\mathrm{TP}_c+\mathrm{FP}_c+\mathrm{FN}_c}
\end{equation}

For the nuScenes, we follow the official evaluation protocol to evaluate the accuracy: nuScenes detection score (NDS) which consists of average translation error (ATE), average scale error (ASE), average orientation error (AOE), average velocity error (AVE), and average attribute error (AAE).

\textbf{Implementation Detail}
In this paper, thanks to the lightweight design of the module, all training is done by a single RTX3090. For the training of the baseline models, we used batches equal to 8 for 80 epochs training and other settings remain as in available implementation.
For the training of MDCNet, we used a batch size of 4 and an Adam optimizer with a learning rate of 3e-5 for the first 15 epochs and 1e-5 for the remaining 25 epochs. The $\alpha$ in Equation 3 is set to 1.2 if the number of pixels is less than 2000, else 1.05. The $\omega_1$, $\omega_2$, and $\lambda_m$ in the loss function are set to 2.0, 2.0, and 1.0 respectively. For the IFST, we trained it with 30 epochs before combining it with MDCNet to get a faster convergence in the early period of training and prevent gradient explosions. 

For the 2D instance segmentation part, we used E2EC\cite{zhang2022e2ec} for KITTI and HTC\cite{chen2019hybrid} for nuScenes to get the best balance between efficiency and accuracy. Considering that the number of other objects is quite limited and some categories of the objects in nuScenes are not available in nuImage, we only report the results of pedestrians and cars detection.  

When we designed the network, we considered different settings in the self-attention Layer. We found that in our network, a sigmoid activation function outperforms the Softmax in the original design\cite{46201} and the combination of Conv-LayerNorm instead of Linear projection in the calculation of K, Q, V accelerates network convergence. So in IFST and MDCNet, the mentioned design is used to replace all self-attention operations. However, for the cross attention, the Softmax function is kept unchanged. Specific comparisons will be shown in the ablation study section. 

\vspace{-1em}
\subsection{Main results}

In Table \ref{ResultKitti} and \ref{ResultNu}, we combine the proposed method with most of the available code baseline models and compare them with the SOTA solution. Here we didn't use any GT sampling when training the baseline model with WYSIWYD, however, for a more convincing comparison, we still remain in this step when retraining the baseline models themselves. 

On the KITTI side, compared with the baseline detector, the proposed MDCNet and IFST provide improvements from 1.29\% to 10.4\% in 3D detection. For Voxel-RCNN, a 12.2\% percent improvement in BEV detection is observed over the original performance. Furthermore, when combined with WYSIWYD, Voxel-RCNN achieves the SOTA pedestrian 3D detection model and surpasses all previous best models by 1.48\%, 2.45\%, 2.97\%. For all other models, our method also brings significant

\begin{table}[H]
	\centering
	\setlength{\tabcolsep}{0.75mm}{
		\begin{tabular}{cccccccc}
			\toprule[0.25mm]
			\multirow{2}{*}{\centering Baselines} & 
			\multirow{2}{*}{\centering \makecell{Completion \\ Method}} & \multicolumn{3}{c}{Car 3D $AP_{R40}$} & 
			\multicolumn{3}{c}{Ped. 3D $AP_{R40}$} \cr \cmidrule[0.25mm](lr){3-5} \cmidrule[0.25mm](lr){6-8}
			& &Easy&Mod.&Hard&Easy&Mod.&Hard \cr
			\cmidrule[0.25mm](lr){1-8}
			\multirow{5}{*}{PV-RCNN\cite{shi2020pv}}& Ori & 92.02 & 84.52 & 82.45 &67.52&60.41&55.23\\
			& SPG\cite{xu2021spg} & +0.43 & +0.95 & +0.34 & +2.35 & +2.14 & +2.47 \\
			& BTC\cite{xu2022behind} & +1.88 & +0.77 & +1.12 & - &- &-\\
			& UYI\cite{zhang2023use} & \textbf{+1.90} &+1.12 & +0.68 & +3.45 & +0.91 & +0.18\\
			& Ours & +0.43 & \textbf{+1.25} & \textbf{+0.89} & \textbf{+7.91} & \textbf{+6.16} & \textbf{+6.70}\\
			
			\cmidrule[0.25mm](lr){1-8} 
			\multirow{5}{*}{Part-$A^2$\cite{shi2020points}}& Ori & 91.72 & 83.08 & 80.45 &67.32 & 60.32&54.49\\
			& SPG$^*$\cite{xu2021spg} & - & - & - &- &- &-\\
			& BTC\cite{xu2022behind} & +0.05 & +1.02 & +0.66 & - &- &-\\
			& UYI\cite{zhang2023use} & +0.23 & \textbf{+1.98} & +0.77 & +0.53 & +0.71 & +0.78\\
			& Ours & \textbf{+0.92} & +0.68 & \textbf{+0.82} & \textbf{+1.29} & \textbf{+3.93} & \textbf{+4.67}\\
			
			\cmidrule[0.25mm](lr){1-8} 
			\multirow{5}{*}{PointPillars\cite{lang2019pointpillars}} & Ori & 87.75 & 78.39 & 75.18 & 57.30 & 51.41 &46.87\\
			& SPG\cite{xu2021spg} & +2.02 & \textbf{+2.97} & +3.67 &+2.35 &+2.14 &+2.47\\
			& BTC\cite{xu2022behind} & +1.66 & +2.77 & +1.03 & - & - &-\\
			& UYI\cite{zhang2023use} & +0.30 & +0.40 & +0.35 &+0.30 &+2.21 &+2.26\\
			& Ours & \textbf{+2.89} & +2.33 & \textbf{+4.91} & \textbf{+4.62} & \textbf{+5.09} & \textbf{+8.77}\\
			\bottomrule[0.25mm]
	\end{tabular}}
	\vspace{2mm}
	\caption{Comparison of the performance improvement by completion methods proposed recently with our solution. The IOU Thresholds are 0.7 and 0.5 for cars and pedestrians. The highest improvement marked in bold. $*$: SPG paper did not provide data on PartA2 and its code is not available.}
	\label{DifferentComp}
\end{table}

\noindent improvements in pedestain detection. Under the 0.7 IOU threshold, we also observe a SOTA performance in Car BEV detection when testing Part-$A^2$. On the nuScenes side, a wide performance improvement is also witnessed on nuScenes performance. Specifically, when combine VoxelNext with the WYIWYD augmented points, we obtain 3.2\% and 2\% improvement in MAP and NDS, which makes this baseline model exceed the latest BEV perception methods.

We also compare the proposed solution with the previous SOTA detector-independent lidar completion methods in Table \ref{DifferentComp} to demonstrate more salient properties of our method.  Specifically, we directly refer to the data in SPG\cite{xu2021spg} and UYI\cite{zhang2023use}, while for BTC\cite{xu2022behind}, we used the completed point cloud output from the completion network as the input to different baseline detector models.  The BtcDet only released configuration on car detection training, therefore the pedestrian items are not reported in the Table. Our proposed method also provides the highest performance improvement in all 3 pedestrian detection categories and most car detection categories when compared to the evaluated methods.

\section{Ablation study}
In this section, we will first show the overall analysis of the proposed model and then assess the effectiveness verification for the different modules. Finally, a qualitative analysis was performed which included visualization of the predicted mesh in 3D space and detection result comparisons. 

\subsection{Overall Analysis}
\begin{figure*}[htp]
	
	\centering
	\subfloat[Result for WYSIWYD augmented car detection]{
		\includegraphics[width=0.95\textwidth]{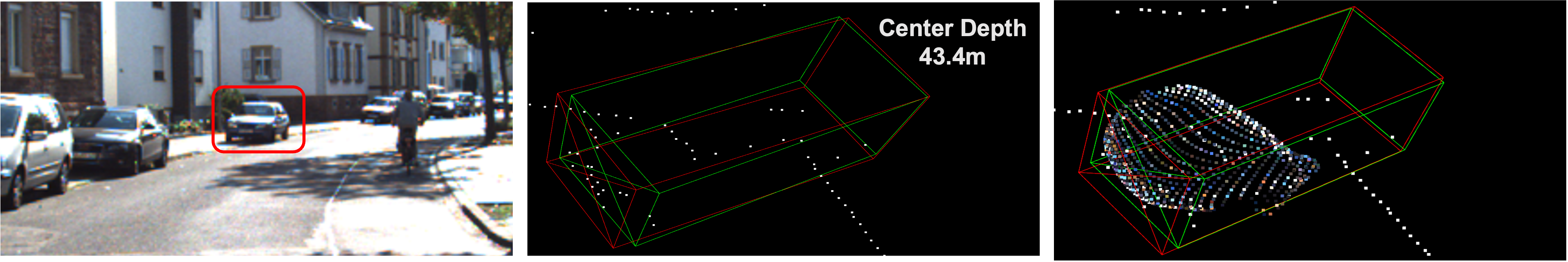}
	}
	\quad
	\subfloat[Improvement for distant car miss detection ]{
		\includegraphics[width=0.95\textwidth]{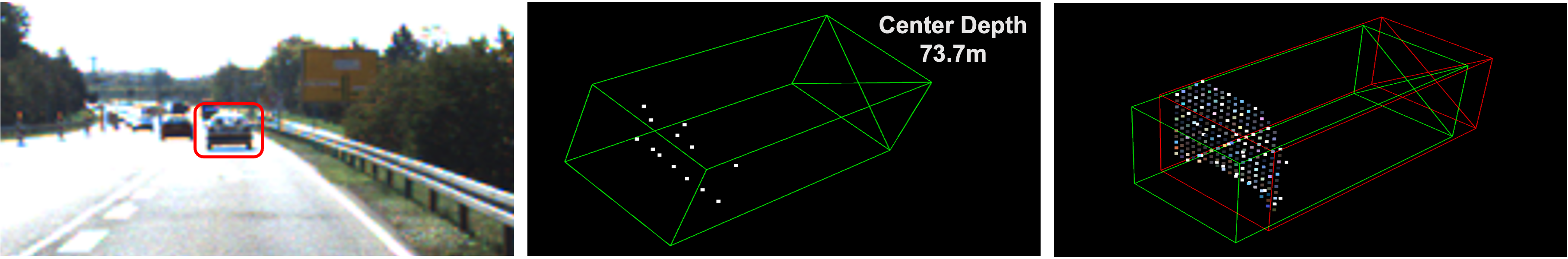}
	}
	\quad
	\subfloat[Result for pedestrian detection: Angle correction and Missing detection restoration]{
		\includegraphics[width=0.24\textwidth]{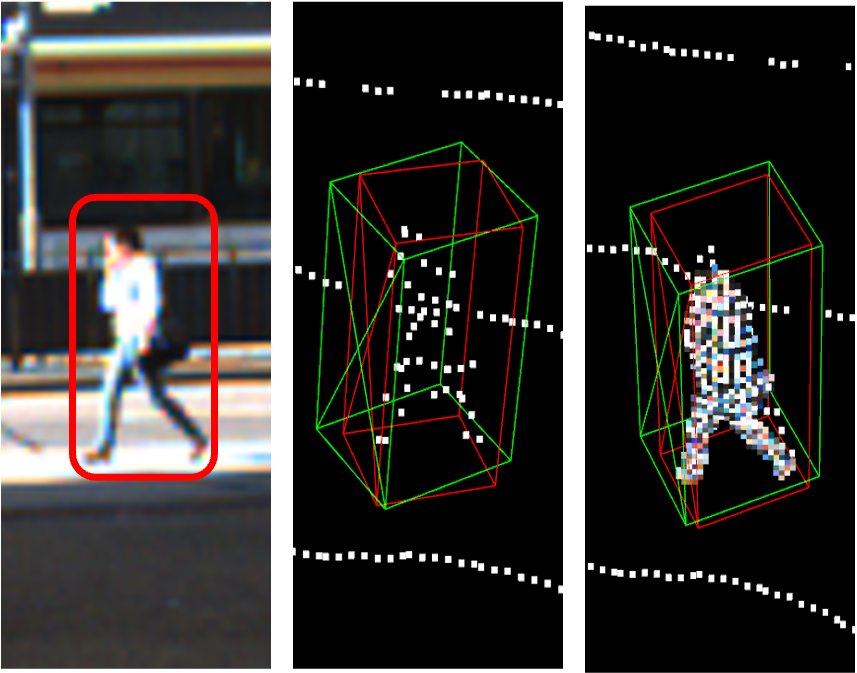}
		\includegraphics[width=0.24\textwidth]{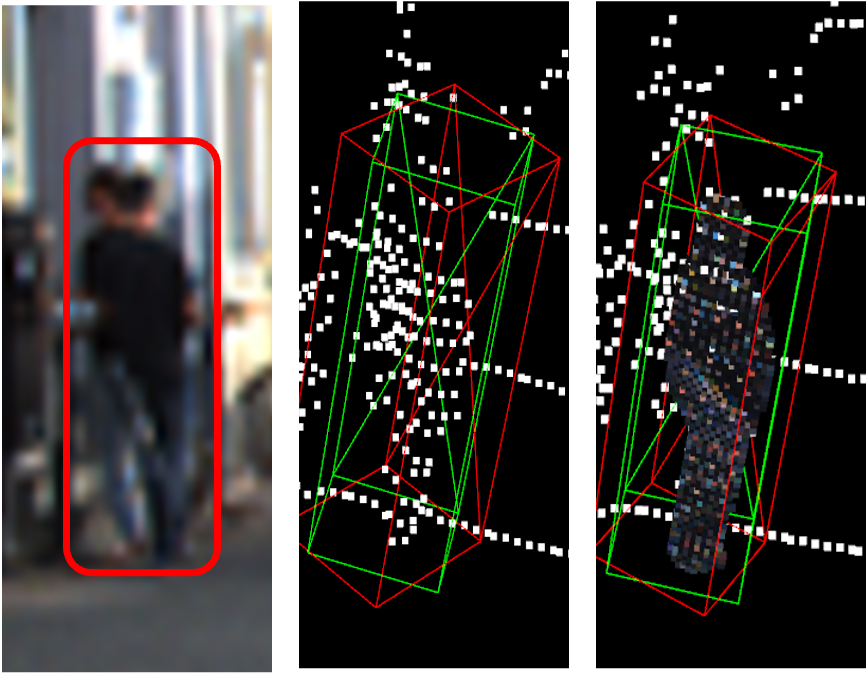}
		\includegraphics[width=0.24\textwidth]{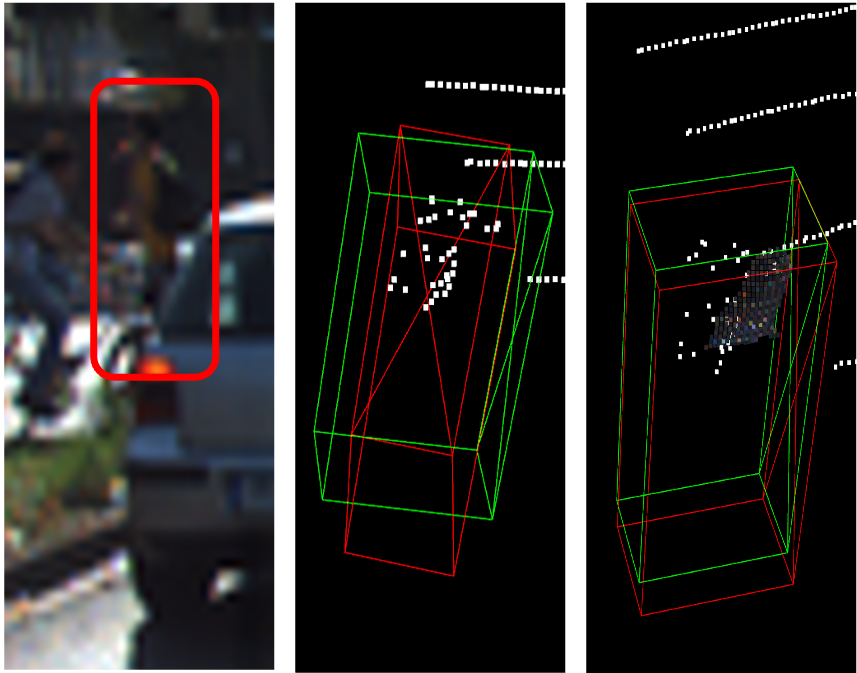}
		\includegraphics[width=0.25\textwidth]{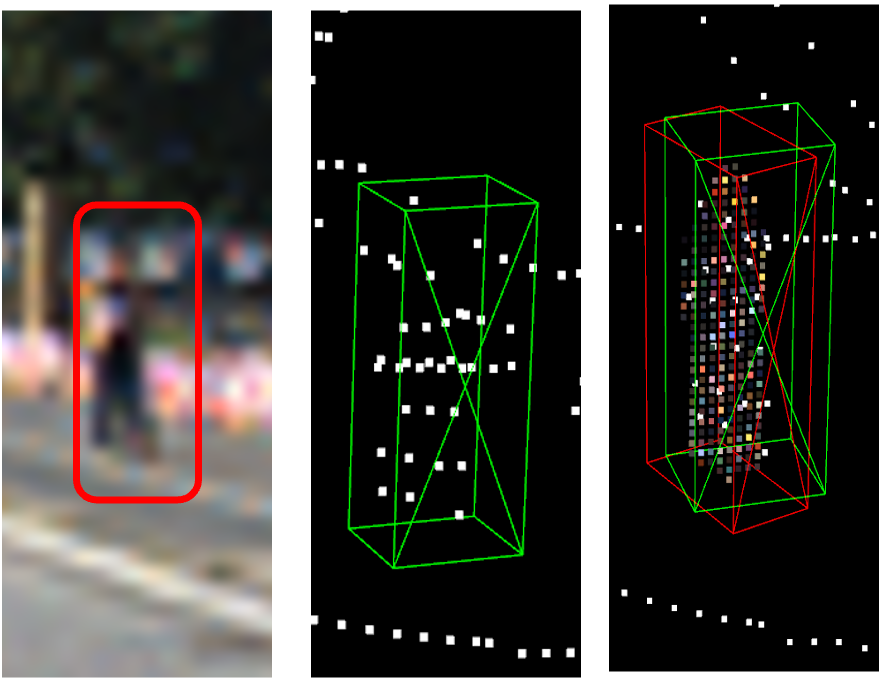}
	}

	\caption{Qualitative verification for effectiveness of the proposed method. For each set of three images, the object positions in the RGB image, the detection results of the original method, and the WYSIWYD augmented point cloud detection results are shown from left to right. 3D Ground truth and predicted box are shown in green and red, respectively.}
	\label{QA_vis}
	
\end{figure*}

\textbf{Vehicle 3D Detection Analysis} In Table \ref{ResultKitti}, we noted the WYSIWYD brings less significant gains in car detection and there is even a decrease in some cases, compared with that of pedestrians. We attribute this to the detector independence of the proposed method. The added complementary point cloud is not perfect, as shown in Fig. \ref{QA_vis}, and from time to time the point cloud boundary exceeds the 3D GT box due to the inaccuracy of the 2D detection. Considering there is no specific design in the downstream detector to filter this noise, this seriously affects the performance of 3D detection under the 0.7/0.7/0.7 thresholds. However, if the thresholds are relaxed to 0.7/0.5/0.5, as shown in Table \ref{AP0.5}, the combination of Voxel-RCNN+WYSIWYD remains optimal in terms of performance. So in this way, the utilization of the proposed model can essentially mitigate the miss detection and low IOU detection problems.

\begin{table}[H]
	\centering
	\setlength{\tabcolsep}{4.5mm}{
		\begin{tabular}{cccc}
			\toprule[0.25mm]
			\multirow{2}{*}{\centering Models} & 
			\multicolumn{3}{c}{Car 3D $AP_{R40}$} \cr 
			\cmidrule[0.25mm](lr){2-4} 
			& Easy&Mod.&Hard \cr
			\cmidrule[0.25mm](lr){1-4}
			PV-RCNN+WYSIWYD & 98.89 & 97.49 & 95.45\\
			Part-A$^2$+WYSIWYD & 98.82 & 95.62 & 95.15\\
			Voxel-RCNN+WYSIWYD & 99.03 & \textbf{97.92} & \textbf{95.68}\\
			\cmidrule[0.25mm](lr){1-4}
			VFF+PV-RCNN& 98.51 & 96.51 & 94.41 \\
			LoGoNet& 98.48 & 96.50 & 94.44 \\
			SFDNet & \textbf{99.32} & 97.04 & 95.01 \\
			\bottomrule[0.25mm]
	\end{tabular}}
	\vspace{2mm}
	\caption{The Car 3D detection AP comparison under 0.7/0.5/0.5 thresholds. To provide a fair comparison, we report the best result in the Moderate categort of all models in the last 20 epochs}
	\label{AP0.5}
\end{table}

As we mentioned in the previous section, since the cross-modality restrain, we cancel the GT-sampling in training process. However, this strategy in fact plays an important role in preventing overfitting. Here we compare the original baseline with the one without the GT sampling augmentation, to further illustrate the improvment bring by WYSIWYD. In Table \ref{aa}, a more obvious improvements can be observed in both pedestain or car detection

\begin{table}[H]
	\centering
	\setlength{\tabcolsep}{0.9mm}{
		\begin{tabular}{ccccccccc}
			\toprule[0.25mm]
			\multirow{2}{*}{\centering \makecell{Baselines}} & 
			\multirow{2}{*}{\centering \makecell{Category}} &
			\multirow{2}{*}{\centering \makecell{Aug.}} & 
			\multicolumn{3}{c}{\centering \makecell{Car 3D $AP_{R40}$}} & 
			\multicolumn{3}{c}{\centering \makecell{BEV 3D $AP_{R40}$}} \cr
			\cmidrule[0.25mm](lr){4-6} \cmidrule[0.25mm](lr){7-9}
			& & & Easy&Mod.&Hard& Easy&Mod.&Hard  \\
			\cmidrule[0.25mm](lr){1-9} 
			\multirow{7}{*}{VoxelRCNN} & \multirow{3}{*}{Car} & \ding{55} & 92.14 & 83.00 & 80.72 & 95.14 & 89.29 & 88.97\\
			& & G & 92.75&85.30&82.94&95.80&91.35&88.99 \\
			& & W & 92.60&\textbf{85.84}&\textbf{83.43}&95.64&\textbf{91.82}&\textbf{89.58} \\
			
			\cmidrule[0.25mm](lr){2-9} 
			& \multirow{3}{*}{Ped.} & \ding{55} & 65.21 & 57.87 & 54.23 & 69.62 & 63.77 & 58.23\\
			& & G & 66.88 & 59.94 & 54.16 & 69.62 & 63.02 & 58.02 \\
			& & W & \textbf{75.56} & \textbf{69.38} & \textbf{64.56} & \textbf{80.10} & \textbf{75.25} & \textbf{68.70} \\

			\bottomrule[0.25mm]
	\end{tabular}}
	\vspace{2mm}
	\caption{Performance comparsion among basline with/without GT-sampling and with the proposed methods. G: with GT-sampling, W: with WYSIWYD}
	\label{aa}
\end{table}

\textbf{Inference Speed comparison} Another feature that deserves to be pointed out is the real-time nature of our algorithm. In Table \ref{infer} we compare its inference time with the previous best-performing models on KITTI. In \cite{wu2022sparse, wu2023virtual}, the time reported is not CUDA synchronized, which means the next frame can actually be processed when the GPU is available, as mentioned in \cite{EPImple}. When this is taken into account, the mentioned method will need more than 200 ms for single-frame inference. However, in our model, thanks to the fact that MDCNet is designed to only complete foreground points, compared to VirConv, the proposed method brings a 34.4 \% efficiency improvement. 

\begin{table}[H]
	
	\centering
	\setlength{\tabcolsep}{1.6mm}{
		\begin{tabular}{c|ccc|c}
			\toprule[0.25mm]
			\multirow{2}{*}{\centering \makecell{Modules}} & 
			\multirow{2}{*}{\centering \makecell{Inference \\ Speed(ms)}} & 
			\multirow{2}{*}{\centering \makecell{Global \\ Completion}} & 
			\multirow{2}{*}{\centering \makecell{Completion Time \\ (if global)}} & \multirow{2}{*}{\centering \makecell{Overall \\ Time(ms)}} \\
			
			& & & &  \\
			\cmidrule[0.25mm](lr){1-5} 
			SFDNet & 66 & \checkmark & 161 & 227 \\
			VirConv & 60 & \checkmark & 161 & 221 \\
			Ours & 145 & \ding{55} & N/A & 145 \\
			
			\bottomrule[0.25mm]
			
	\end{tabular}}
	\vspace{2mm}
	\caption{Comparision of inference time in pseudo-points based completion. PENet\cite{hu2021penet} is used as completion network as mentioned in SFDNet and VirConv paper. The Voxel-RCNN is used as the baseline.}
	\label{infer}
\end{table}

Specifically, the reported 145ms in the Table is composed of the forward propagation time of Voxel-RCNN 44ms, the forward propagation time of E2EC 43ms and the 58ms of WYSIWYD.

\textbf{Conditional Analysis} In addition, in order to explore in which scenarios our proposed method brings greater improvements, we analyzed the performance gain on Voxel-RCNN and PV-RCNN using distance and occlusion degree as indicators. As shown in Table \ref{occ_distance_ana}, 
\begin{table}[H]
	\centering
	\setlength{\tabcolsep}{0.4mm}{
		\begin{tabular}{ccccccccc}
			\toprule[0.25mm]
			\multirow{2}{*}{\centering \makecell{Baselines}} & 
			\multirow{2}{*}{\centering \makecell{Category}} &
			\multirow{2}{*}{\centering \makecell{With \\ WYSIWYD}} & 
			\multicolumn{3}{c}{\centering \makecell{Distance(m)}} & 
			\multicolumn{3}{c}{\centering \makecell{Occlusion}} \cr
			\cmidrule[0.25mm](lr){4-6} \cmidrule[0.25mm](lr){7-9}
			& & & 0-20&20-40 &40-Inf& 0 & 1 & 2  \\
			\cmidrule[0.25mm](lr){1-9} 
			\multirow{8}{*}{VoxelRCNN} & \multirow{2}{*}{Ped.} & \ding{55} & 62.11 & 34.36 & 1.04 & 60.14 & 19.39 & 4.59\\
			& & \checkmark & 68.26 & 51.42 & 7.46 & 70.82 & 34.01 & 10.39 \\
			\cmidrule[0.25mm](lr){2-9} 
			& \multicolumn{2}{c}{\textbf{Improvement}} & +6.15 & +17.06 & +6.42 & +10.68 & +14.62 & +5.80 \\
			
			\cmidrule[0.25mm](lr){2-9} 
			& \multirow{2}{*}{Car} & \ding{55} & 90.78 & 78.19 & 28.95 & 74.37 & 69.90 & 55.11\\
			& & \checkmark & 91.03 & 78.70 & 36.12 & 76.49 & 70.89 & 54.47 \\
			\cmidrule[0.25mm](lr){2-9} 
			& \multicolumn{2}{c}{\textbf{Improvement}} & +0.25 & +0.52 & +7.17 & +2.12 & +0.99 & -0.37 \\

			\cmidrule[0.25mm](lr){1-9}
			
			\multirow{8}{*}{PV-RCNN} & \multirow{2}{*}{Ped.} & \ding{55} & 60.81 & 30.94 & 0.62 & 55.68 & 15.23 & 4.56\\
			& & \checkmark & 65.77 & 50.36 & 12.21 & 67.54 & 30.22 & 9.42 \\
			\cmidrule[0.25mm](lr){2-9} 
			& \multicolumn{2}{c}{\textbf{Improvement}} & +4.96 & +19.42 & +11.59 & +11.86 & +14.99 & +4.86 \\
			
			\cmidrule[0.25mm](lr){2-9} 
			& \multirow{2}{*}{Car} & \ding{55} & 90.43 & 77.63 & 29.69 & 73.79 & 69.04 & 53.22\\
			& & \checkmark & 90.48 & 78.85 & 38.32 & 77.65 & 70.21 & 54.07 \\
			\cmidrule[0.25mm](lr){2-9} 
			& \multicolumn{2}{c}{\textbf{Improvement}} & +0.05 & +1.22 & +8.63 & +3.64 & +1.17 & +0.85 \\
			\bottomrule[0.25mm]
	\end{tabular}}
	\vspace{2mm}
	\caption{3D detection performance at different level of distance and different occlusion. The thresholds for IOU and other settings are same as mentioned in Table 2.}
	\label{occ_distance_ana}
\end{table}
As shown in this Table, our approach has a substantial improvement for the detection of distant objects: we can bring up to 19.42\% performance improvement for objects in the range of 20-40 meters. Even for targets at more than 40 meters away from the camera, we achieve at least 6.42\% performance improvement. 

\subsection{Component-wise Analysis}
To further explore the detail of the performance of our method, we split the detection results into 3 bins by distance and different occlusions as marked by KITTI. The results are shown in Table \ref{ideal}. In addition to this, we also compared the inference time between the proposed completion method and the previous pseudo-point-based solution.

\begin{table*}\label{ideal}
	\centering
	\setlength{\tabcolsep}{1.5mm}{
		\begin{tabular}{c|ccc|c|cccccccccccc}
			\toprule[0.25mm]
			\multirow{2}{*}{\centering \makecell{Modules}} & 
			\multirow{2}{*}{\centering \makecell{Cluster \\ Aggregation}} & 
			\multirow{2}{*}{\centering \makecell{KDE \\ Feature}} &
			\multirow{2}{*}{\centering \makecell{Label}} & 
			\multirow{2}{*}{\centering \makecell{Mesh \\ MSE Loss \\ }} &
			\multicolumn{3}{c}{Car 3D $AP_{R40}$} & 
			\multicolumn{3}{c}{Car BEV $AP_{R40}$} &
			\multicolumn{3}{c}{Ped. 3D $AP_{R40}$} &
			\multicolumn{3}{c}{Ped. BEV $AP_{R40}$} \cr
			\cmidrule[0.25mm](lr){6-8} \cmidrule[0.25mm](lr){9-11} \cmidrule[0.25mm](lr){12-14} \cmidrule[0.25mm](lr){15-17} & & & & & Easy&Mod.&Hard&Easy&Mod.&Hard &Easy&Mod.&Hard &Easy&Mod.&Hard \\
			\cmidrule[0.25mm](lr){1-17} 
			WYSIWYD &  &  & N/A & 489.3 & 91.12 & 82.00 & 82.61 & 92.01 & 88.75 & 89.09 & 63.02 & 61.78 & 56.72 & 72.40 & 68.80 & 63.02 \\
			WYSIWYD & \checkmark &  & N/A & 311.9 & 92.18 & 84.71 & 82.99 & 94.61& 92.10 & 88.72&69.22&63.95&60.76&74.03&70.18&64.65\\
			WYSIWYD & \checkmark & \checkmark & N/A & 283.4 & 92.45 & 85.76 & 83.34 & 95.58& 91.68 & 89.44&70.38&66.57&61.93&75.43&71.02&66.23 \\
			\cmidrule[0.25mm](lr){1-17} 
			 WYSIWYD & \checkmark & \checkmark & 2D & 277.3 & 93.24 & 89.24 & 86.54 & 95.79 & 92.98 & 90.53 & 74.36 & 69.99 & 65.50 & 78.01 & 73.66 & 70.61 \\
			WYSIWYD & \checkmark & \checkmark & 2D+3D & 275.9 & 94.69 & 90.06 & 87.74 & 95.98 & 93.06 & 92.93 & 78.11 & 72.98 & 68.82 & 80.51 & 75.99 & 73.31 \\
			
			\bottomrule[0.25mm]
			
	\end{tabular}}
	\vspace{2mm}
	\caption{Effect of different designs in WYSIWYD on KITTI val object mesh prediction. The downstream 3D detector selected is PV-RCNN in this experiment. The results of using 2D detection and 3D segmentation labels are also reported.}
	\label{ablation_fop}
\end{table*}

\textbf{IFST design verification} Here we decomposed the module in IFTS and verified its effectiveness in Table \ref{ITFS_ver}. By adding of 2D mask size feature, lidar 2D location feature, Local embedding layer, and Sinusoidal Embedding, we get 0.48\%, 2.92\%, 1.10\%, 0.65\% improvement. In summary, compared to the vanilla PointNet++\cite{qi2017pointnet++} which is widely used in intra-frustum segmentation, IFST offers 10.58\% performance improvement in terms of mIOU.  

\begin{table}[H]\label{ITFS_ver}
	\centering
	\setlength{\tabcolsep}{1.4mm}{
		\begin{tabular}{c|cccc|c}
			\toprule[0.25mm]
			\multirow{2}{*}{\centering \makecell{Modules}} & 
			\multirow{2}{*}{\centering \makecell{Mask Size}} & 
			\multirow{2}{*}{\centering \makecell{Lidar 2D \\ location}} & 
			\multirow{2}{*}{\centering \makecell{Neighbour \\ Embedding}} & \multirow{2}{*}{\centering \makecell{Sinusoidal \\ Embedding}} & 
			\multirow{2}{*}{\centering \makecell{mIOU(\%)}} \\
			
			& & & & & \\
			\cmidrule[0.25mm](lr){1-6} 
			PointNet++ & N/A & N/A & N/A & N/A & 78.44 \\
			IFST &  &  &  &  & 83.87 \\
			IFST & \checkmark &  &  &  & 84.35 \\
			IFST & \checkmark & \checkmark &  &  & 87.27 \\
			IFST & \checkmark & \checkmark & \checkmark &  & 88.37 \\
			IFST & \checkmark & \checkmark & \checkmark & \checkmark & 89.02 \\
			\bottomrule[0.25mm]
			
	\end{tabular}}
	\vspace{2mm}
	\caption{Effect of different design in IFST on KITTI val lidar segmentation. For the IFST case, we adopt a vanilla Transformer with mentioned modification in projection layer and activation function}
\end{table}

\textbf{Ablation of MDCNet design} In Table \ref{ablation_fop} In Table \ref{ablation_fop} we show the direct relationship between generated mesh quality and detection result. When adding local-global aggregation and KDE guide, we can obtain 3.76\% and 4.79\% performance increase in PV-RCNN, for car moderate detection and pedestrian moderate detection respectively. In this process, the MSE mesh loss continues to fall from 489.3 to 283.4. 

Considering that our method requires both 2D instance segmentation and 3D points segmentation, we also report the performance of using the two labels directly in the last 2 lines of Table \ref{ablation_fop}. As can be seen, our method still has great potential: a better 2D segmentation is enough to improve the performance by another 4.3\% to 6.4\% percent in terms of moderate 3D detection.

\begin{figure}[htbp] 
	\centering 
	\includegraphics[width=0.5\textwidth]{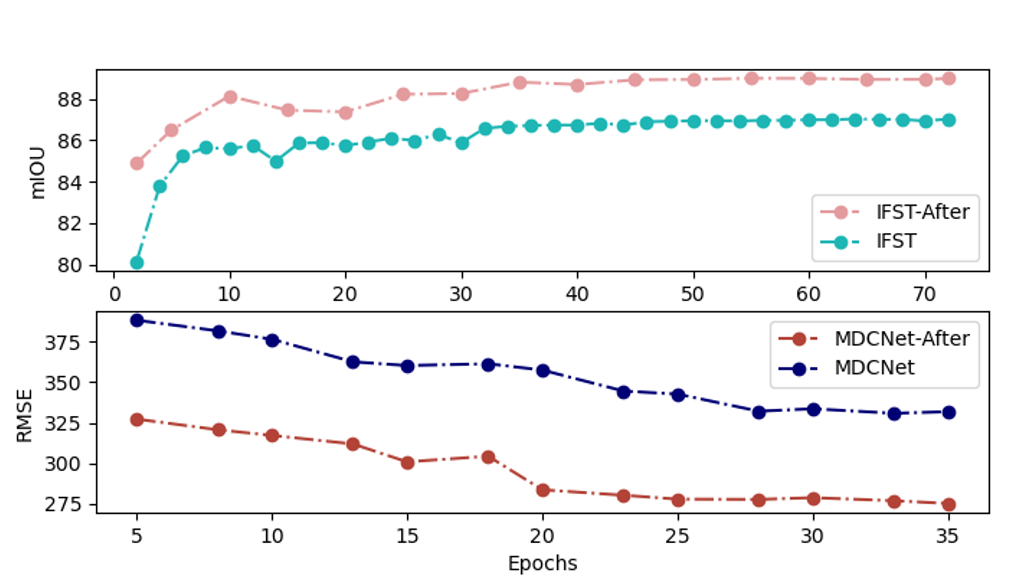} 
	\caption{Comparsion of proposed module with different self-attention layer design and activation function} 
	\label{Converge} 
\end{figure}

\textbf{Transformer Layer design} In the Figure \ref{Converge}, we show the role of normlized projection layer and sigmoid activation function in the convergence rate of IFST and MDCNet. As shown, these design not only accelrate the training, but also improve the final accuracy from 2\% to 9\%.

\subsection{Qualitative Analysis}
In order to illustrate the completion points brought by WYSIWYD more intuitively, we visualize some examples of obvious performance improvements brought by it in Figure \ref{QA_vis}.

For Figure \ref{QA_vis}(a) and Figure \ref{QA_vis}(b), we report a better boundary box estimation in terms of IOU and missing detection. The proposed model recovered the missing details in the visible part in a mesh-deformation manner, which is especially important when the lidar data is extremely sparse, as shown in Figure \ref{QA_vis}(b). Figure \ref{QA_vis}(c) shows the impact of WYSIWYD added points on pedestrian detection. From left to right, in the first 3 sets of images, we observe an improvement in the estimation of the bearing angle. Furthermore, the last 3 images prove that the miss-detection issue can be eliminated by adding points as well. 

\section{Conculsion}
In this work, we proposed a solution to improve the foreground depth in 3D detection in a mesh-deformation manner. In this process, we discard the traditional time-consuming global completion and our final result gets SOTA performance, especially in pedestrian 3D detection. Extensive experiments on baseline models demonstrate the effectiveness and robustness of our proposed model. 

\bibliography{reference}
\bibliographystyle{IEEEtran}


\vspace{11pt}


\vfill

\end{document}